\documentclass[sigconf,nonacm]{acmart}
\AtBeginDocument{%
  }

\setcopyright{acmlicensed}
\copyrightyear{2018}
\acmYear{2018}
\acmDOI{XXXXXXX.XXXXXXX}
\acmConference[Conference acronym 'XX]{Make sure to enter the correct
  conference title from your rights confirmation email}{June 03--05,
  2018}{Woodstock, NY}
\acmISBN{978-1-4503-XXXX-X/2018/06}

\usepackage{microtype}
\usepackage{graphicx}
\usepackage{subcaption}
\usepackage{booktabs} 

\usepackage{hyperref}
\usepackage{amsmath}
\usepackage{mathtools}
\usepackage{amsthm}
\usepackage{enumitem}

\usepackage{xurl}
\usepackage[textsize=tiny]{todonotes}

\usepackage{listings} 
\usepackage{xspace}
\usepackage{cleveref}
\usepackage{graphicx}  
\usepackage{pifont}  
\usepackage{marginnote}
\usepackage{algorithm}
\usepackage{algpseudocode}
\usepackage{multirow}
\usepackage{tcolorbox}
\usepackage[table]{xcolor}

\newcommand{\delegator}{delegator\xspace}

\newcommand{\greenagent}{judge agent\xspace}

\newcommand{\greenagents}{judge agents\xspace}
\newcommand{\Greenagents}{Judge agents\xspace}
\newcommand{\GreenAgents}{Judge Agents\xspace}
\newcommand{\whiteagent}{subject agent\xspace}

\newcommand{\whiteagents}{subject agents\xspace}
\newcommand{\Whiteagents}{Subject agents\xspace}
\newcommand{\WhiteAgents}{Subject Agents\xspace}

\newif\ifappd
\appdfalse

\newif\ifsubmit
\submitfalse

\newcommand{\remove}[1]{}

\newcommand{\reviewer}[3]{
  \expandafter\newcommand\csname #1\endcsname[1]{
    \ifsubmit
      \ignorespaces
    \else
      \textcolor{#3}{[#2: ##1]}
    \fi
  }
}

\reviewer{xiaoyuan}{Xiaoyuan}{orange}

\begin{document}

\title[AgentBeats: Agentifying Agent Assessment for Openness, Standardization, and Reproducibility]{AgentBeats: Agentifying Agent Assessment for Openness, Standardization, and Reproducibility}


\author{
Xiaoyuan Liu$^{1}$,
Jianhong Tu$^{2}$,
Yuqi Chen$^{2}$,
Siyuan Xie$^{1}$,
Sihan Ren$^{1}$,
Tianneng Shi$^{1}$,
Gal Gantar$^{3}$,
Evan Sandoval$^{1}$,
Donghyun Lee$^{1}$,
Daniel Miao$^{4}$,
Peter J. Gilbert$^{5}$,
Nick Hynes$^{5}$,
Mauro Staver$^{5}$,
Warren He$^{5}$,
David Marn$^{5}$,
Andrew Low$^{6}$,
Xi Zhang$^{5}$,
Elron Bandel$^{7}$,
Michal Shmueli-Scheuer$^{7}$,
Siva~Reddy$^{8,9}$,
Alexandre Drouin$^{9,10}$,
Alexandre Lacoste$^{10}$,
Ramayya Krishnan$^{11}$,
Elham Tabassi$^{12}$,
Yu~Su$^{13}$,
Victor~Barres$^{14}$,
Chenguang Wang$^{2}$,
Wenbo Guo$^{15}$,
Dawn Song$^{1}$
}

\renewcommand{\shortauthors}{}

\begin{abstract}
Agent systems are advancing quickly across domains, but their evaluation remains fragmented. Most benchmarks rely on fixed, LLM-centric harnesses that require heavy integration, create test-production mismatch, and limit fair comparison across diverse agent designs. The root problem is the lack of an open, agent-agnostic assessment interface. We advocate Agentified Agent Assessment (AAA), where evaluation is performed by \greenagents and all participants interact through standardized protocols: A2A for task management and MCP for tool access. Conventional benchmarking defines two separate interfaces, one for the benchmark and one for the agent, while AAA only needs one; this yields a generic, unified framework that separates assessment logic from agent implementation and enables reproducible, interoperable, and multi-agent evaluation.

We further introduce AgentBeats as a concrete realization of AAA: we identify five practical operation modes that make standardized assessment compatible with real-world constraints on openness, privacy, and reproducibility; and we provide recommended practices that let both agent developers and benchmark designers adopt AAA with minimal effort, turning evaluation from ad-hoc integration into a reusable, production-aligned process. 

To evaluate our design at scale, we conduct two studies: a five-month open competition that drew 298 \greenagents across 12 categories together with 467 \whiteagents from independent participants, showing that AAA applies across a heterogeneous range of benchmarks; and a case study on coding agents that confirms agentified evaluation preserves fidelity with the public record while surfacing previously missing head-to-head results, yielding research insights about agent design. Combining a community-scale field study and a controlled coding case study, we verify that AAA delivers coverage, practicality, and fidelity across heterogeneous scenarios at scale. Together, AAA and AgentBeats offer a clear path toward open, standardized, and reproducible agent assessment.
\end{abstract}

\maketitle

\begingroup
\renewcommand{\thefootnote}{}
\footnotetext{
$^1$ UC Berkeley;
$^2$ University of California, Santa Cruz;
$^3$ EPFL;
$^4$ The Harker School;
$^5$ RDI Foundation;
$^6$ Independent;
$^7$ IBM Research;
$^8$ Mila;
$^9$ McGill University;
$^{10}$ ServiceNow Research;
$^{11}$ Carnegie Mellon University;
$^{12}$ The Brookings Institution;
$^{13}$ The Ohio State University;
$^{14}$ Mercor;
$^{15}$ University of California, Santa Barbara.
}
\endgroup

\section{Introduction}

Large language model (LLM)-based agent systems have seen rapid growth in recent years and are increasingly deployed across diverse tasks. Examples include coding agents~\cite{anthropic2026claudecode,openai2026codex} that autonomously create and iteratively refine software repositories, browser-based agents~\cite{perplexitycomet,openaichatgptatlas} that retrieve information and execute web actions, and computer-use agents~\cite{sager2025comprehensive} that interact directly with graphical interfaces via keyboard and mouse control. This evolution marks a clear shift from standalone foundation models toward agentic systems that combine LLMs with tools, memory, and execution harnesses. Such integration substantially expands the scope of tasks that can be performed and enables capabilities that are difficult or impossible for isolated models.

\begin{figure}[t]
    \centering
    \includegraphics[width=0.7\linewidth]{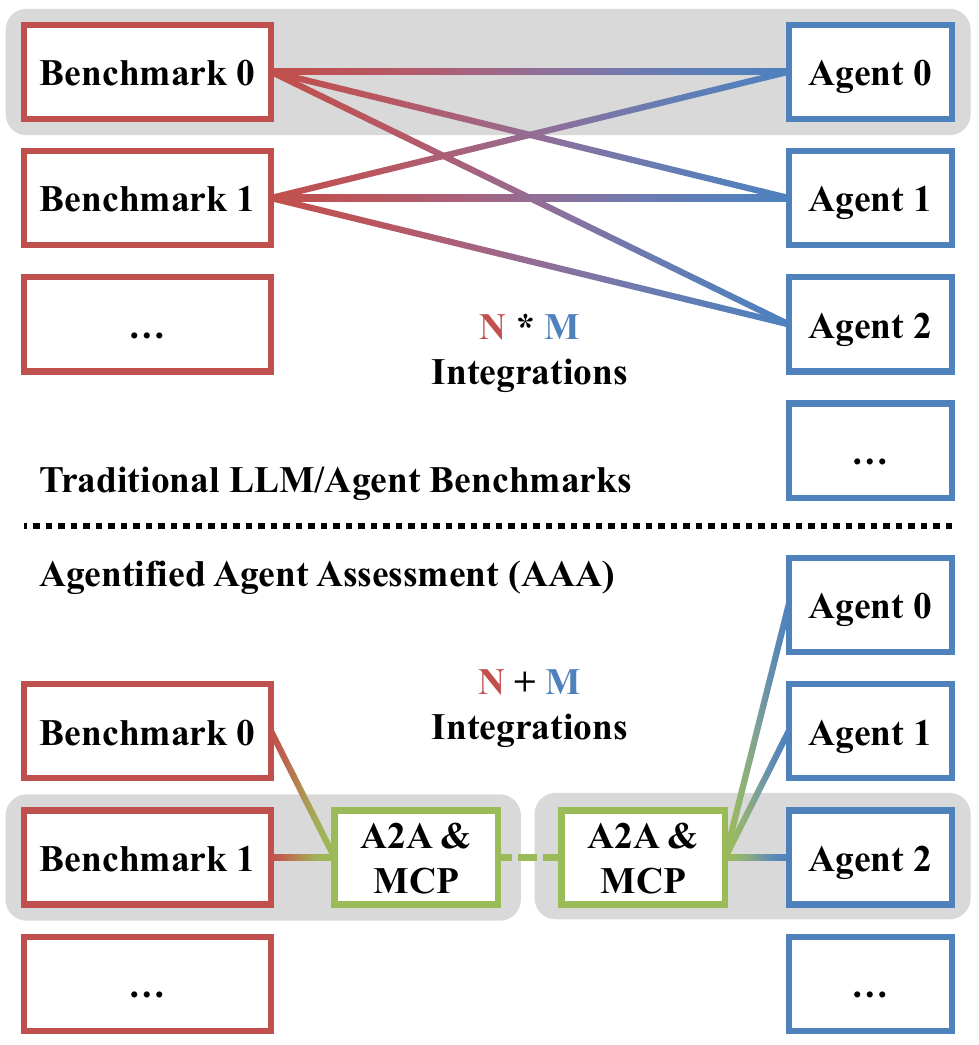}
    \caption{Comparison between Traditional LLM/Agent benchmarks and AAA. AAA reduces the number of integrations from $N \times M$ to $N + M$, while completely separating the benchmark and target agent as shown in the gray boxes.}
    \label{fig:aaanm}
    \vspace{-0.5cm}
\end{figure} 

As a result, a wide range of evaluations and benchmarks have been proposed to measure different aspects of agent capabilities. However, a fundamental compatibility problem quickly emerges. Agent systems differ widely in their architectures and interfaces, while benchmarks impose their own assumptions about input formats, tool APIs, and environment control~\cite{bandel2026readyforgeneral}. An arbitrary agent cannot be directly evaluated on an arbitrary benchmark. In practice, this mismatch leads to a fragile and inefficient evaluation workflow. Benchmark developers typically integrate a small set of well-known agents through custom, benchmark-specific implementations. This integration effort is often substantial, yet the resulting agent coverage remains limited. As a concrete real-world example, agent implementations such as \citet{openhandsbenchmarks} maintain a dedicated benchmark directory that aggregates dozens of external benchmarks, requiring nontrivial, benchmark-specific adaptation and customization.

The lack of standardization has severe consequences. First, it requires significant effort to align benchmark interfaces with target agents. Even after investing time to learn each benchmark and agent codebase, evaluating M agents on N benchmarks essentially requires N * M bespoke integration efforts as shown in Figure~\ref{fig:aaanm}.
Second, some integrations require changes to both agents and benchmarks. Integrations developed by different parties may also follow different evaluation logic, which may lead to inconsistent or unfair results~\cite{bandel2026generalagentevaluation}. 
Lastly, evaluating a production agent often requires additional customization, creating a test-production mismatch. As a result, reported metrics may fail to capture behaviors or risks that only emerge in real-world deployment.

\begin{table*}[ht]
  \small
  \begin{tabular}{lll}
    \toprule
    & \textbf{Traditional Benchmarking} & \textbf{Agentified Agent Assessment (AAA)} \\
    \midrule
    \textbf{Benchmark-agent relation} & Tightly coupled & Complete separation \\
    \textbf{Eval target} & Primarily LLMs with fixed harnesses & Any agent conforming to the A2A protocol \\
    \textbf{Eval general-purpose agent} & Needs benchmark-specific customizations & Allows one general agent to work with multiple assessments \\
    \textbf{Interface} & Benchmark-specific and implementation-dependent & Standardized: A2A for task management, MCP for tool access \\
    \textbf{Realism} & Prone to test--production mismatch & Directly reflects production-level performance \\
    \textbf{Multi-agent support} & Difficult; requires bespoke integrations & Natively supported through standardized interfaces \\
    \bottomrule
  \end{tabular}
  \caption{Comparison between traditional agent benchmarking and the proposed Agentified Agent Assessment (AAA) paradigm.}
  \label{tab:compare}
  \vspace{-0.3cm}
\end{table*}

Standardization is key to resolving this compatibility problem. To enable a benchmark to directly evaluate any agent, the benchmark must make no assumptions about the agent’s internal structure or prior knowledge. However, agents expose diverse interfaces and benchmarks target a wide range of capabilities, which makes it challenging to define a standard that simultaneously (1) is flexible enough to support diverse evaluation scenarios and (2) is simple and lightweight enough to encourage broad adoption.
This tension motivates the following questions:

\begin{itemize}
    \item Can we define a standardized, agent-agnostic assessment interface that makes minimal assumptions about agent internals, remains interoperable across benchmarks and agents, and reduces test–production mismatch?
    \item Further, can such a standard be expressive and lightweight enough to support a broad spectrum of evaluation styles and deployment constraints, and what limitations are unavoidable in practice?
\end{itemize}


Administering an agent benchmark, which involves environment setup, tool provisioning, user simulation, and outcome judging, is itself a well-defined agentic task. This observation motivates treating the benchmark as an agent.
In this paper, we introduce \emph{Agentified Agent Assessment (AAA)}, a paradigm that treats benchmarks themselves as agents and enforces a clean separation between assessment logic and the agents being evaluated. Under AAA, all interactions are mediated through existing, production-facing protocols, enabling benchmarks and agents to interoperate without bespoke integration. 
The protocol combination is flexible enough to support benchmarks for most known agent types, such as terminal-use, computer-use, coding, and web agents.
Table~\ref{tab:compare} summarizes the key differences between traditional agent benchmarking and the AAA paradigm we propose. AAA addresses the limitations of existing approaches through standardization and decoupling, and naturally enables multi-agent evaluation due to its support for agent communications. 

We then present \emph{AgentBeats}, a practical system that instantiates AAA and demonstrates how agentified assessment can be deployed under real-world constraints, including openness, privacy, and reproducibility. AgentBeats supports both single-agent and multi-agent evaluations, accommodates heterogeneous agent implementations, and provides multiple operation modes that align evaluation workflows with how agents are actually developed and deployed. Through the design of AgentBeats, we argue that agentifying agent assessment requires minimal additional effort while yielding substantial benefits. 

In principle, this resolves the compatibility problem between agents and benchmarks that we set out to address; whether the paradigm holds up against the diversity of real-world agents and benchmarks, however, calls for validation at scale.

Therefore, we put the paradigm to the test through two complementary studies. The first is a five-month open competition that invited hundreds of external developer teams; the second is a focused case study evaluating four representative coding agents on three coding benchmarks through a single AAA-standardized pipeline. Concretely, we ask whether the paradigm holds up along three dimensions:

\begin{itemize}
    \item \textbf{Coverage.} Can arbitrary, heterogeneous benchmarks be converted into \greenagents? Simple tool-use benchmarks are straightforward to agentify, but it is less clear whether the same approach extends to more complex and varied evaluation scenarios.
    \item \textbf{Practicality.} Is the conversion practical and easy to adopt, and does it actually improve interoperability between agents and benchmarks in real use?
    \item \textbf{Fidelity.} Does agentifying a benchmark change its measured outcomes? Differences in prompts and evaluation flow are known to shift agent performance; does agentification introduce such shifts, and how severe are they?
\end{itemize}

Our studies answer each question in turn. On coverage, the competition drew 298 \greenagents across 12 categories together with 467 \whiteagents, agentifying dozens of existing benchmarks that span coding, web browsing, healthcare, and multi-agent games. On practicality, these agents came from independent developers in multiple programming languages and frequently declared their evaluation logic through natural-language prompts, signaling low adoption friction. On fidelity, our results match the public record where comparable, and the standardized setup produces useful research insights the public record cannot, such as a co-adaptation effect we verify between models and their native harnesses. Together, these results show that agentifying agent assessment is practical at scale and a viable path toward open, standardized, and reproducible evaluation.

\section{Agentified Agent Assessment Paradigm}
\label{sec:aaa}

\subsection{Motivation}

AAA is motivated by two key statements that guide the design of a universal agent assessment paradigm~\cite{bandel2026position}. 
First, \textbf{a universal agent benchmark must be self-contained and present tasks in a way that a human can understand and complete without prior knowledge or external information channels}.  Most agents are designed to operate under human instruction, and benchmarks should reflect this interaction pattern by providing clear, self-explanatory instructions and environments. In doing so, benchmarks simulate a human operator issuing commands and judging outcomes, while enabling broader coverage and larger-scale evaluation. It is important to note that this principle does not conflict with presenting tasks in an LLM-friendly format. Specifying required output formats or including few-shot examples remains a recommended practice for high-quality benchmarks. In fact, this aligns naturally with the self-contained task principle: humans also benefit from explicit formats for fair evaluation, and few-shot examples help clarify task requirements.

Second, \textbf{one of the most effective ways to encourage adoption is to reuse existing, widely adopted agent operation standards and repurpose them for evaluation}. Specifically, by leveraging the A2A protocol~\cite{a2a2025specification2025} for task management and the MCP~\cite{mcp2025specification} for tool access, we can cover most agent evaluation scenarios. A2A provides a clear mechanism for task specification, progress tracking, multimodal data exchange, and bidirectional communication. MCP, in contrast, offers structured, self-contained, and format-enforced access to tools and functions. 
Both protocols are already widely adopted and are not inherently limited to any particular task domain or modality. Most modern agents support A2A natively or can be adapted via a lightweight wrapper using the official A2A libraries, while MCP has become a common practice for environment interaction and external resource access. These specific protocols are not the only possible choice; other protocols could in principle serve the same role. What matters is that the chosen protocol is already widely used, which keeps integration effort low and makes community adoption realistic.

Based on these two statements, we further advocate that \textbf{agent evaluation itself should be agentified}, since evaluating other agents is a well-defined agentic task. This design brings several advantages.

\begin{enumerate}
    \item First, it enables natural adoption of evaluation standards. Conventionally, standardizing a benchmark requires specifying two separate interfaces: one for the benchmark to expose its environment, instructions, and scoring logic, and one for the agent to receive instructions and emit actions. Framing evaluation as one agent assessing another reduces these to a single agent interface applied to both sides, which aligns communication with the agent-to-agent paradigm and fits cleanly within the A2A protocol. Task specifications become standard A2A task messages, and tested agents can interact with environments through MCP tool calls without requiring benchmark-specific harnesses.
    \item Second, agentification turns the benchmark's own internal components, such as user simulators and LLM-as-judge scorers, into agentic sub-components that share the same paradigm as the outer assessment. Beyond simply separating simulation logic from the tested agent, this lets these components be composed, reused across benchmarks, and developed independently, and they can even be exposed as standalone A2A services rather than hard-coded inside a single harness.
    \item Third, benchmark design and result judgment become flexible. Agentifying a benchmark amounts to \emph{internalizing} its evaluation workflow inside the \greenagent, which can be done in two ways. \emph{Programmatic internalization} hard-codes the evaluation logic in the \greenagent's implementation, reproducing the original benchmark's behavior precisely. \emph{Semantic internalization} instead expresses the evaluation logic as natural-language instructions, which reduces coding effort, handles complex evaluation more flexibly, and natively supports LLM-as-a-judge scoring.
\end{enumerate}

Based on such observations, we design Agentified Agent Assessment (AAA) as a new paradigm that treats benchmarks themselves as agents to address the universality challenge in agent evaluation. 

\begin{figure}[hbt]
    \centering
    \includegraphics[width=0.98\linewidth]{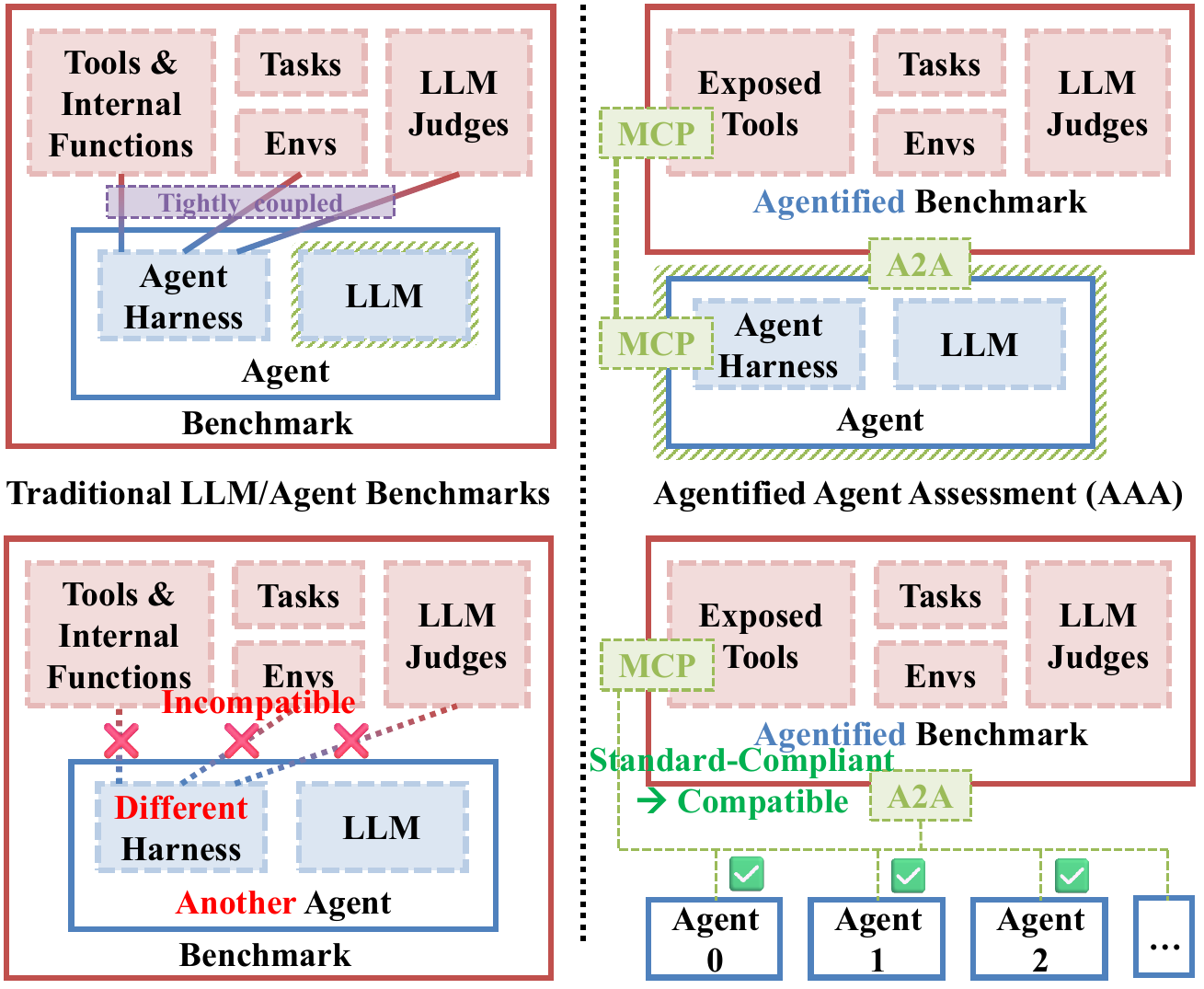}
    \caption{Complete separation between benchmarks and agents. Traditional benchmarks are tightly coupled with fixed agent harness and typically allow only LLM replacement as illustrated by the green shaded area, whereas agentified benchmarks allow the entire agent to be swapped.}
    \label{fig:couple}
\end{figure} 

As mentioned in Table~\ref{tab:compare}, the core difference between traditional agent assessment and AAA is the complete separation between benchmarks and the agents being evaluated. In practice, this means that the evaluated agent can reside in a separate repository and be developed by an independent party. This separation relies on two requirements in AAA benchmark design, following the earlier statements. First, the benchmark must make no assumptions about the agent’s internal structure or prior knowledge, assuming only compatibility with A2A and MCP. Second, tasks must be described in a fully self-contained manner through A2A messages, with all necessary context and tools explicitly provided.

As a result, unlike traditional assessments that typically allow only LLM replacement or support a small set of predefined agent harnesses, as shown in Figure \ref{fig:couple}, AAA enables any A2A- and MCP-compatible agent to be evaluated without additional integration effort. This reduces the integration cost from N * M agent-benchmark combinations to N + M protocol-level integrations as shown in Figure~\ref{fig:aaanm}.

Besides reducing integration cost, AAA can also potentially improve evaluation efficiency via conducting adaptive assessments. Because the \greenagent owns the full evaluation session, the set of tasks need not be enumerated in advance: the \greenagent can generate them adaptively, probe weaknesses, or stop early. AAA thus reframes a benchmark from a fixed dataset paired with a scoring function into an interactive evaluation process, concentrating effort on informative cases and avoiding redundant ones. For example, when an agent fails early test cases, the \greenagent can skip later ones likely to yield the same result; when an agent does well, it can instead generate harder cases to better differentiate performance.

Beyond efficiency, this standardization also enables interoperability and reproducibility. An unopinionated open standard with recommended practices encourages the development of interoperable benchmarks and agents, while an evaluation management platform with multiple operation modes supports reproducible assessments across diverse use cases and deployment settings.

\subsection{Workflow}

Based on these principles, we now describe the concrete workflow and participant roles under AAA.
AAA is a standardized evaluation paradigm in which a benchmark is realized as an agent that assesses other agents through agent-native interactions, enabling plug-and-play interoperability, clear separation of assessment-side logic, and broad flexibility in how tasks are specified and outcomes are judged.
Under this paradigm, an agent assessment involves three participant roles: a \delegator, a \greenagent (or assessor agent), and one or more \whiteagents (or assessee agents).

\begin{itemize}
    \item The \textit{\greenagent}, acting as an agentified benchmark, implements the evaluation logic and judging procedures, including the test dataset, evaluation workflow, and the metrics to be reported. Despite the name, a \greenagent is distinct from an LLM-as-a-judge: the latter, when used at all, is merely one optional scoring component inside a \greenagent.
    \item There may be one or more \textit{\whiteagents}. A \whiteagent is the evaluation target that understands A2A tasks and can connect to an MCP server. In multi-agent assessments, multiple A2A-compatible agents are evaluated as separate targets, and their performance metrics are reported independently.
    \item The \textit{\delegator} selects a benchmark by choosing a \greenagent, describes the evaluation requirements, and specifies the targets by selecting one or more \whiteagents.
\end{itemize}

To better understand these roles, one can imagine a car owner bringing a vehicle to an auto repair shop. The car owner (the \delegator) chooses a repair shop or a maintenance package offered by different technicians (\greenagents) to inspect the car (\whiteagent). The technicians performing the examination are responsible for producing the final report.

In practice, the \delegator may be a script or a service that issues A2A tasks, while both the \greenagent and \whiteagents must run A2A services. The AAA paradigm itself does not prescribe how roles are mapped to real-world actors or how they are deployed. In Section \ref{sec:ab}, we describe how this paradigm is instantiated in AgentBeats through five operation modes to accommodate real-world requirements.

\begin{figure}[t]
    \centering
    \includegraphics[width=0.75\linewidth]{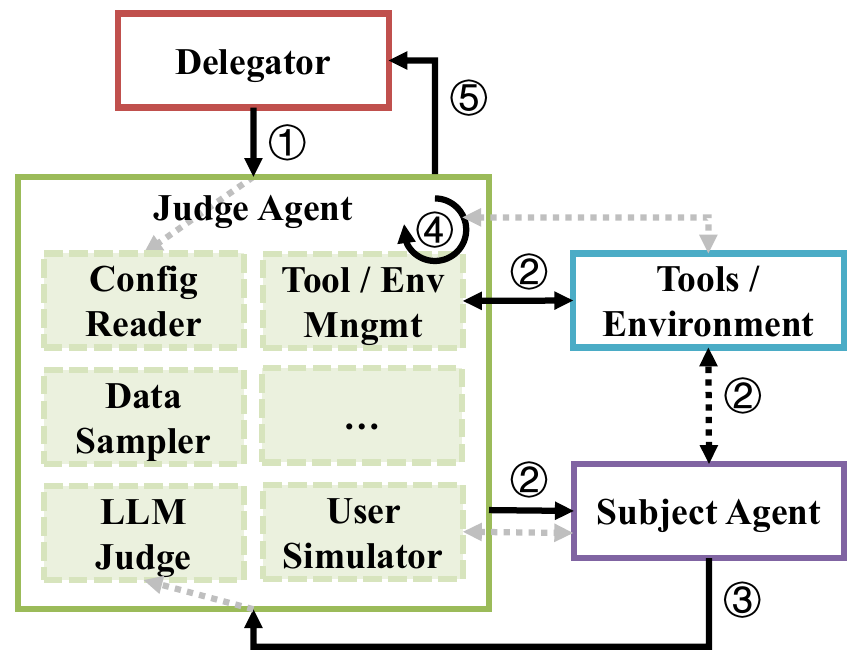}
    \caption{AAA interaction pattern. The internal structure of the \greenagent is illustrative rather than prescriptive. Section~\ref{sec:rec} discusses the recommended practices for designing \greenagents.}
    \label{fig:steps}
    \vspace{-0.5cm}
\end{figure} 

Figure \ref{fig:steps} illustrates the interaction pattern defined by AAA.

\begin{enumerate}
    \item \textbf{Evaluation request delegation}. The \delegator identifies the evaluation goal and configuration, selects the appropriate \greenagent and \whiteagents, specifies the desired result format, and sends a kick-off request to the \greenagent via A2A containing this information. AAA does not enforce a specific request format, and we provide a reference format in the discussion of the AgentBeats system.
    \item \textbf{Task instruction distribution and access authorization}. Based on the request details, the \greenagent retrieves relevant data and prepares the evaluation environment, such as launching or configuring an MCP server. It then distributes task instructions, along with environment access information, to the \whiteagents.
    \item \textbf{Task fulfillment}. Upon receiving the task instructions, \whiteagents attempt to complete the task within the allotted time. During execution, depending on task requirements, \whiteagents may interact with the \greenagent one or more times, invoke MCP tools, explore the environment, and take actions.
    \item \textbf{Performance scoring}. After receiving task completion notifications from \whiteagents, or when a timeout occurs, the \greenagent revokes environment access and inspects both the \whiteagents’ responses and the final environment state to assess performance. Each \whiteagent is scored according to predefined metrics.
    \item \textbf{Result reporting}. The \greenagent reports evaluation results back to the \delegator in the requested format.
\end{enumerate}

This structure naturally extends to multi-agent assessments. In single-agent benchmarking, the \greenagent does not need to consider participant roles or capacities, as there is only one evaluation target. In multi-agent assessments, \greenagents define, during the design phase, the expected roles and capacities of the \whiteagents they can assess. Depending on the task type, these specifications may vary. In general, multi-agent tasks can be categorized as collaborative or adversarial.

In adversarial multi-agent tasks, such as chess games~\cite{kaggle2025gamearena}, werewolf-style games~\cite{xu2023language}, or simulated CTF competitions~\cite{zou2026ctfagent}, the roles of \whiteagents are typically well defined, and specifications can directly follow the team structure or winning conditions of the task.
On the other hand, in collaborative multi-agent tasks, the arrangement depends on whether the task explicitly defines a division of labor among agents. Developers may choose either to (1) treat different agents as independent \whiteagents and score them accordingly, or (2) when the multi-agent structure primarily serves to improve performance while the task itself can be viewed as single-agent, one can organize auxiliary agents as sub-agents and conduct the evaluation as a single-agent assessment.

\section{Related Works}

\begin{table*}[tb]
\centering
\small
\renewcommand{\arraystretch}{1.25}
\resizebox{\linewidth}{!}{
\begin{tabular}{@{}>{\raggedright\arraybackslash}p{2.8cm}>{\raggedright\arraybackslash}p{3.0cm}>{\raggedright\arraybackslash}p{3.0cm}>{\raggedright\arraybackslash}p{3.0cm}>{\raggedright\arraybackslash}p{3.0cm}>{\raggedright\arraybackslash}p{3.0cm}>{\raggedright\arraybackslash}p{3.0cm}@{}}
\toprule
\textbf{Dimension} & \textbf{NeMo \cite{nemoevaluator}} & \textbf{HAL \cite{hal}} & \textbf{Harbor \cite{harborframeworkteam2026harborframework}} & \textbf{CUBE \cite{lacoste2026cubestandardunifyingagent}} & \textbf{Exgentic \cite{bandel2026generalagentevaluation}} & \textbf{AgentBeats (AAA)} \\
\midrule
Coverage strategy & Increase harness number for a single agent system & Define new evaluation interface & Leverage terminal as interface standard & Define new benchmark infrastructure interface & Define new evaluation protocol & Leverage A2A and MCP as interface standard \\
Standard / interface used & \textbf{New:} task-specific harness collection inside one repo (no shared agent protocol) & \textbf{New:} minimal evaluation-oriented agent interface & \textbf{Ecosystem-backed:} terminal I/O + checking scripts as shared structure & \textbf{Ecosystem-backed:} MCP for tool access; Gym-style API (reset/step/evaluate) over JSON-RPC & \textbf{New:} Unified Protocol (task/context/actions); adaptors bridge A2A, MCP, CLI, tool-calling & \textbf{Ecosystem-backed:} A2A for task management + MCP for tool access \\
Required engineering effort (to reach broad coverage) & \textbf{High:} many harnesses must be built for coverage & \textbf{Moderate:} adaptation needed to adopt the new interface & \textbf{Low:} only need instructions + scripts & \textbf{Moderate:} adaptation needed to adopt the CUBE protocol & \textbf{Moderate:} per-benchmark decomposition into task/context/actions required & \textbf{Low:} only need to support A2A and MCP \\
Scalability via community adoption & \textbf{Hard:} typically requires sustained, centralized industry effort to curate and maintain large harness collections & \textbf{Hard:} adoption friction due to introducing a new standard & \textbf{Good potential:} builds on already widely used shell script format & \textbf{Moderate:} partial reuse of existing protocols & \textbf{Moderate:} new protocol derived from existing patterns; adaptor-based integration & \textbf{Good potential:} builds on already widely adopted A2A/MCP standards \\
Fit for UI-centric tasks (e.g., browser interaction) & \textbf{Supports:} harnesses can target diverse modalities & \textbf{Supports:} the new standard does not limit multimodal evaluations & \textbf{Weaker:} terminal serialization less suitable for UI tasks & \textbf{Supports:} infrastructure layer is task-type agnostic & \textbf{Supports:} Unified Protocol agnostic to modality; agents retain native UI mechanisms & \textbf{Supports:} multimodal A2A exchange cover UI cases \\
Risk of test-production mismatch & \textbf{High:} benchmark-specific integrations diverge from production agents & \textbf{High:} eval-only interface risks mismatch & \textbf{Moderate:} well aligned for terminal-native tasks (e.g., coding agents) & \textbf{N/A:} depends on the evaluation framework & \textbf{Low:} agents preserve native protocols (incl. A2A/MCP); mediation is benchmark-side only & \textbf{Low:} built directly on production-ready A2A/MCP protocols \\
\bottomrule
\end{tabular}
}
\caption{Comparison of representative related works.}
\label{tab:related}
\vspace{-0.6cm}
\end{table*}

While we advocate agentifying agent assessment, other approaches also help mitigate the agent-benchmark compatibility challenge. We group them by their overall strategy.

\textit{Per-benchmark first-party harnesses.} NeMo Evaluator~\cite{nemoevaluator} takes a monorepo approach, building one task-specific harness per benchmark inside a single curated repository. The strength of this design is alignment: each harness can be hand-tuned to its target benchmark, so a model-only evaluation extracts as much as possible from any given setup. The downside is engineering effort: every new benchmark requires a new first-party harness, and integrating a third-party agent that does not match the curated harness still requires bespoke adaptation.

\textit{Scenario-agnostic lightweight standards.} A second line introduces a new lightweight standard that any benchmark and agent can implement, trading per-benchmark curation effort for breadth. HAL~\cite{hal} specifies a minimal functional interface and adds cost tracking plus a leaderboard for a more holistic view of agent performance. Because that interface is designed for evaluation rather than deployment, agents tested through it can diverge from the version actually shipped to users, reintroducing a test-production mismatch. CUBE~\cite{lacoste2026cubestandardunifyingagent} focuses on the deployment side instead, standardizing how benchmark environments are packaged and spawned so the same task can run under different harnesses. Exgentic~\cite{bandel2026generalagentevaluation} introduces a Unified Protocol of task, context, and actions, paired with non-intrusive adaptors that preserve agents' native protocols (CLI, MCP, A2A, and tool-calling), though adopting it on the benchmark side requires decomposing each benchmark into that structure. These standards significantly reduce per-benchmark migration effort and enable reuses within the ecosystem, but coverage then depends on broad adoption of the new standard, which is itself a challenge.

\textit{Scenario-specific lightweight standards.} A related strand narrows the standardization target to a single domain. BrowserGym~\cite{dechezelles2025browsergymecosystemwebagent} exemplifies this in the web setting, consolidating nine web-agent benchmarks (e.g. WebArena~\cite{zhou2024webarena}, WebLINX~\cite{lu2024weblinx}) behind a single Gym-style observation/action space so an agent implemented once runs across all of them. BALROG~\cite{paglieri2024balrog} is a similar example in the game domain, aggregating existing reinforcement-learning environments into a unified benchmark for evaluating LLM and VLM agents on long-horizon interactive tasks. Such efforts work well when the domain is well-defined and the interaction pattern is shared, but each new domain needs its own standard, and benchmarks that span domains do not naturally fit any one of them.

\textit{Reusing existing protocols.} A third line builds on protocols already adopted. Harbor~\cite{harborframeworkteam2026harborframework} builds its standardization on top of the terminal: stdin/stdout for I/O, the shell binaries for tool use, and an interactive shell for the environment. This drastically lowers the integration burden and is especially natural for agents that already operate a bash shell. The limitation comes from this choice itself: tasks that do not project naturally onto a terminal, such as those involving multimodal artifacts, require additional, non-standard conventions layered on top of the terminal. AAA sits in this same group but reuses A2A and MCP for the agent-benchmark interface, leaving the remaining diversity to community-contributed \greenagents rather than to a new evaluation-specific abstraction. Table~\ref{tab:related} summarizes the comparison with representative works.

Beyond standardizing the interaction between benchmarks and agents, a complementary line of work focuses on standardizing how environments expose functionality to agents. Rather than addressing benchmark-agent compatibility, these efforts seek to replace or augment human-oriented interfaces with agent-native abstractions that are more efficient, structured, and easier for autonomous systems to interact with.

\textit{Agent-native interfaces.}
Many interfaces today are designed primarily for humans. 
For example, the dominant interface to the Web is through HTTP and HTML websites. 
A complementary direction is to explore agent-native interfaces.
\citet{lu2025build} argue that human-facing websites, including visual pages, DOM trees, screenshots, and ad hoc developer APIs, are inefficient for autonomous agents. 
They propose Agentic Web Interfaces (AWIs): agent-native web interfaces with structured state, explicit action spaces, and safety boundaries.
This direction is orthogonal to AAA: AWIs define what a web environment exposes to agents, whereas AAA defines how subject agents, judge agents, and tools communicate during assessment.

\section{Realizing AAA: AgentBeats}
\label{sec:ab}

Applying AAA to real-world scenarios requires more concrete system designs. In practice, one must account for different assessment participants, such as individual developers or organizations, and their roles in a complete evaluation workflow. While the use of A2A and MCP in AAA addresses standardization, three additional dimensions must be considered: openness, privacy, and reproducibility.
\textit{Openness} implies encouraging modularized, open-source implementations of both agents and benchmarks, supporting collaborative iteration and publicly accessible assessment services.
\textit{Privacy} requires support for closed-source LLMs and agents. In some cases, to ensure fair evaluation and avoid data leakage or overfitting, it is desirable not only to evaluate on hidden data, but also to support private evaluation workflows through closed-source \greenagents.
\textit{Reproducibility} calls for clear specification and automation of assessment procedures, stability or determinism in \greenagent behavior, and explicit state management for \whiteagents.

To address diverse real-world development and deployment scenarios, we design five operation modes. These modes differ in participant-role mapping, deployment requirements, \delegator assignment, and result presentation.
For example, some assessments target open-source agent implementations specified as GitHub repositories or Docker images, while others evaluate already deployed, closed-source A2A services. From a system perspective, the agent input to an assessment may take the form of a Git URL, a Docker image reference, or an A2A endpoint URL. We refer to the former cases, where a Git URL or Docker image is provided, as agent blueprints, and to the latter as agent instances. Depending on how agents are provided and by whom, different operation modes apply.
To support a complete assessment workflow, each operation mode follows the same three-stage lifecycle, with minor variations in specific steps. We first describe this shared lifecycle, then discuss the design details and differences among operation modes. Finally, we address scalability-related considerations.

\subsection{Assessment Lifecycle}

\begin{figure}[h]
    \centering
    \includegraphics[width=0.9\linewidth]{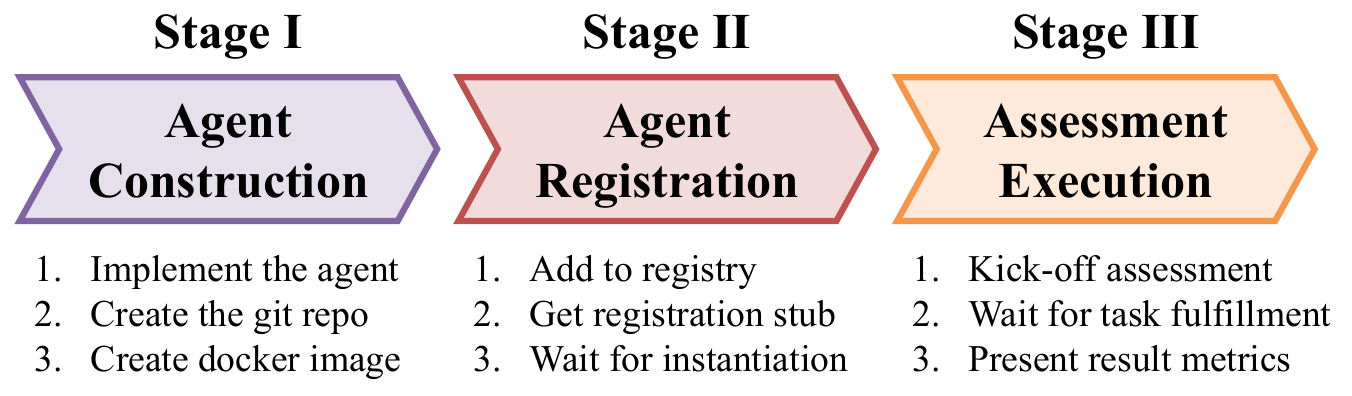}
    \caption{Assessment Lifecycle in AgentBeats. Some steps may be optional in certain operation modes.}
    \label{fig:stages}
\end{figure}

Figure \ref{fig:stages} illustrates a complete assessment lifecycle with three stages: agent construction, agent registration, and assessment execution.

\textit{Stage I. Agent construction.} The goal of this stage is to prepare either an agent blueprint or a live A2A service. Since the \delegator’s role is limited to sending A2A messages, which can often be implemented with a few lines of script, we focus primarily on the effort required for agent construction. After implementing the agent, developers may choose to open-source the agent by creating a publicly accessible Git repository. Additionally, by providing a Dockerfile, developers can expose an easily reproducible A2A service. A recommended practice is to make the LLM API key and A2A host configurable through Docker environment variables, further improving reproducibility.

\textit{Stage II. Agent registration.} This stage gathers all agents relevant to an assessment and optionally makes them publicly accessible. Registration can be as simple as placing agent codebases under a shared directory or specifying agent service URLs in a script. The key objective is to abstract each agent into an easy-to-reference identifier for later stages. In some operation modes, such identifiers are returned as stubs from the registration procedure. When agent blueprints are provided, an optional instantiation step prepares the corresponding live agent instances. The next stage can begin only after all instantiations are complete.

\textit{Stage III. Assessment execution.} An assigned \delegator initiates the evaluation by following the interaction pattern defined in the AAA paradigm. After task completion, results are either returned to the \delegator who requested the assessment or made publicly available to support openness.

Next, we further concretize the details of each stage by examining different operation modes. Some steps in certain modes may be optional, simplified, or omitted altogether.

\subsection{Operation Modes}

\begin{table*}[t!]
\centering
\resizebox{\linewidth}{!}{%
\begin{tabular}{|l|lllll|}
\hline
\textbf{} & \textbf{Local Mode} & \textbf{Remote Mode} & \textbf{Hosted Mode} & \textbf{Proxy Mode} & \textbf{CI Mode} \\ \hline
\textbf{Participants} & One local developer & \begin{tabular}[c]{@{}l@{}}Remote developer(s)\\ + AB platform\end{tabular} & \begin{tabular}[c]{@{}l@{}}Remote developer(s)\\ + AB platform\end{tabular} & \begin{tabular}[c]{@{}l@{}}One local developer\\ + optional remote developer(s)\\ + AB platform\end{tabular} & \begin{tabular}[c]{@{}l@{}}Remote developer(s)\\ + CI platform\end{tabular} \\ \hline
\textbf{\begin{tabular}[c]{@{}l@{}}Where are agents \\ instantiated\end{tabular}} & Local machine & Remote servers & \begin{tabular}[c]{@{}l@{}}AB platform-\\ managed servers\end{tabular} & \begin{tabular}[c]{@{}l@{}}Local machine \\ + Remote servers\end{tabular} & \begin{tabular}[c]{@{}l@{}}CI platform-\\ managed servers\end{tabular} \\ \hline
\textbf{\begin{tabular}[c]{@{}l@{}}Who kicks off \\ the assessment\end{tabular}} & A script on local machine & AB platform & AB platform & AB platform & \begin{tabular}[c]{@{}l@{}}CI platform \\ (e.g. push triggers)\end{tabular} \\ \hline
\textbf{\begin{tabular}[c]{@{}l@{}}Where are \\ the results displayed\end{tabular}} & Local terminal & AB platform & AB platform & AB platform & \begin{tabular}[c]{@{}l@{}}CI platform \\ (via committed files)\end{tabular} \\ \hline
\textbf{Best suited for} & \begin{tabular}[c]{@{}l@{}}Local development\\ or on-prem  assessments\end{tabular} & \begin{tabular}[c]{@{}l@{}}Public assessments \\ of private agents\end{tabular} & \begin{tabular}[c]{@{}l@{}}Public assessments using \\ public agent blueprints\end{tabular} & \begin{tabular}[c]{@{}l@{}}Local development\\ with remote participants\end{tabular} & \begin{tabular}[c]{@{}l@{}}Auditable assessments \\ using public agent blueprints\end{tabular} \\ \hline
\end{tabular}%
}
\caption{\centering AgentBeats operation modes. The AB platform denotes a centralized platform that orchestrates AAA following the design described in Section \ref{sec:ab}. The CI platform denotes continuous integration services, such as GitHub Actions.}
\label{tab:modes}
\vspace{-0.7cm}
\end{table*}

We provide five operation modes that all satisfy the AAA paradigm but differ in their real-world use cases in Table \ref{tab:modes}. A modularized AgentBeats system exposes different system components to support these modes, requiring no or minimal changes to agent implementations while switching between these modes. Below, we describe each operation mode in terms of its suitable scenarios, participant-role mapping, concretized lifecycle, and required system components.

\textbf{Local Mode}. As the simplest and most straightforward realization of the AAA paradigm, local mode requires all participating agents to be instantiated as local services on the same host. This mode is typically suited for local development or fully private, on-prem evaluations. In this setting, all AAA participant roles are mapped to one local developer, with the \delegator implemented as a lightweight script.

A common pattern is to implement or download all agent codebases into separate directories during stage I, then write a script that launches each agent service during stage II. When executing stage III, the script starts or restarts all agent services on predetermined ports, sends the kick-off A2A message, and finally prints the results to the terminal or writes them to a result file. Local mode does not require any dedicated platform components; the only requirement is that all agents conform to the A2A protocol.

\textbf{Remote Mode}. A natural way to orchestrate evaluation and encourage openness is to introduce a centralized platform that manages agents and assessment resources. Remote mode, hosted mode, and proxy mode all adopt this approach and rely on a centralized service component. A platform provider is responsible for deploying and maintaining this backend service and making it accessible to agent developers and assessment users.

In remote mode, during stage I, agent providers deploy their own A2A services and make them publicly accessible or reachable by the platform backend. They then submit agent instance information, such as service URLs, to the platform for registration. In stage III, a platform user can request an assessment by selecting registered agents, then monitor execution progress and results through the platform. As implied by its name, agents in remote mode are deployed on remote servers and remain private to other platform users. This mode is well suited for public, collaborative assessments of potentially closed-source agents.

Remote mode relies on a centralized AgentBeats platform that provides agent registration, kick-off message dispatching, and result tracking. It also requires a concrete kick-off request format, since the platform may manage multiple \greenagents with different benchmark functionalities. Such a request typically includes three components: a prompt describing the evaluation goal, the URLs of \whiteagents with role assignments (e.g., black and white in a chess match), and a customizable configuration provided by the requester.

\textbf{Hosted Mode}. Also built on a centralized platform, hosted mode offers a more streamlined and user-friendly experience by allowing agent providers to submit agent blueprints instead of live service URLs. During registration in stage II, an agent provider can submit a GitHub repository that follows a predefined standard (e.g., buildpacks) or a Docker image. The platform then instantiates the agent and verifies that its A2A service is running correctly. Once verified, the agent is marked as ready for subsequent assessments.

Beyond the capabilities required for remote mode, hosted mode requires the platform to handle agent instantiation and service health checks. Instantiation can be implemented using platform-managed clusters or by leveraging third-party container-based serverless infrastructure for improved scalability. As an additional advantage, whereas remote mode typically supports only a single instance per registered agent, hosted mode allows the platform to provision multiple instances to support large-scale evaluations. 

\textbf{Proxy Mode}. To further support agent development, proxy mode provides an option that combines local development with platform-based evaluation. Consider a developer building a new agent for an upcoming assessment, where other participating agents are already registered on the platform. Remote mode requires full deployment, while hosted mode may incur long build times. Proxy mode addresses this gap by allowing developers to connect a temporary local development server to the platform and reverse-proxy communications between local and remote agents.

From a lifecycle perspective, proxy mode does not require a Git repository or image build. Instead, it assumes an A2A-compatible development server is running locally. In stage II, the local agent connects to the AgentBeats platform to obtain a temporary platform identifier and a message tunnel. By requesting an assessment using this temporary identifier together with other registered agents, and optionally streaming platform-side logs back to the local terminal, developers can test their code against existing agents without leaving their local environment.
Supporting proxy mode requires the platform to provide reverse-proxy functionality and expose execution logs through APIs. A platform-provided command-line tool or SDK that automates tunnel management is also important for a smooth developer experience. With proxy mode, \greenagent developers can directly test benchmark workflows against well-known agents, while \whiteagent developers can validate their implementations against agentified benchmarks without downloading datasets or preparing environments.

\textbf{Continuous Integration (CI) Mode}. Reliance on a centralized platform may raise concerns about trust and auditability. Providing an option that avoids a black-box service improves transparency and reproducibility. Motivated by this consideration, CI mode leverages existing public CI infrastructure, such as GitHub Actions, to realize the AAA paradigm without a dedicated platform. This mode can be viewed as migrating local-mode evaluation into CI pipelines.

There are several ways to automate agentified assessments in CI workflows. A simple approach is to commit all files required for local-mode execution into a repository and configure triggers that invoke the evaluation script on each new commit. In this setup, stage I corresponds to pushing all relevant agent codebases into the repository, along with CI configuration for dependency installation. During stage II, LLM API keys, host addresses, and other environment variables are provided through the CI platform’s secret and token management features to facilitate agent instantiation. Each commit triggers stage III, and results are available in CI logs or committed to a separate branch as artifacts.

Although CI mode does not depend on a centralized platform, it benefits from well-documented reference implementations. Providing example repositories with CI workflows as templates helps establish standard practices. To further encourage modular development, each agent can be wrapped as a subdirectory or submodule with its own Dockerfile or configuration. The CI workflow then orchestrates containerized agent services and executes the assessment in a fully automated and reproducible manner.

\section{Developer Recommended Practice}
\label{sec:rec}

AgentBeats helps \whiteagent developers test their agents, and also helps benchmark designers standardize evaluation interfaces, reducing integration effort on both sides. Here we discuss the implications of practicing AAA and leveraging the AgentBeats system. Specifically, we provide recommended practices for standardizing interfaces while remaining flexible, without introducing excessive additional burden. This section also serves as a guide to help organize the development process for both agents and benchmarks.

\subsection{Developing \WhiteAgents}

The basic requirement for \whiteagent developers is to ensure that their final implementation can accept A2A tasks and connect to external MCP servers specified in the A2A message. In general, there are two cases: (1) the developer already has an agent implementation that does not support these protocols but wants to evaluate it using a \greenagent under AAA, and (2) the developer plans to build a new agent from scratch. In the first case, there are three aspects to check and patch: A2A compatibility, MCP tool access support, and environment adaptivity.

\textbf{A2A compatibility}. If an existing agent does not support A2A (e.g., it exposes a custom API) or only partially supports A2A (e.g., it does not support artifact transfer for media inputs), it is recommended to preserve the original agent package and implement a thin external A2A wrapper that translates task instructions. Since most LLM agents accept natural language instructions in some form before acting, this translation is usually feasible. A concrete example is a coding agent that operates through terminal. In such a case, one can implement a wrapper A2A service that forwards task instructions into a terminal with a clean local working directory, then retrieves output files as artifacts after execution and returns them in the A2A response.

\textbf{MCP tool access support}. While many modern agent frameworks support MCP integration, the key challenge in AgentBeats is that the MCP services used by a \whiteagent may be provided by the \greenagent and specified dynamically in the A2A message at runtime. Supporting dynamic MCP configuration and loading new tools often requires additional customization. A simple approach is to handle this within the same thin A2A wrapper layer. In Section~\ref{sec:ext}, we discuss an alternative approach: providing a default MCP server during the setup phase and dynamically configuring its tool list.

\textbf{Environment adaptivity}. Beyond A2A and MCP, some assessments may require additional capabilities that are natural for humans, thus still being self-contained, but difficult for certain agents. For example, a browser-use task may require navigating to a specific URL, while a terminal-use task may require SSH access to a remote machine. To support such assessments, developers should either implement the required workflows explicitly or incorporate internal tools for the needed capabilities.

In the second case, since AAA only enforces communication protocols, the main considerations largely match those of building a general agent: which framework to use, which LLM to choose, what harness pattern to adopt (e.g., CoT~\cite{wei2022chain} or ReAct~\cite{yao2022react}), and whether to incorporate a multi-agent structure, etc. We note that, in practice, it is often effective to optimize against a single \greenagent (and potentially a small set of peer \whiteagents) first, using local mode or proxy mode. This makes development more purpose-driven and efficient. Meanwhile, we encourage developers to make instruction processing as robust as possible. For example, extracting URLs from browser-use instructions is often more reliable when done via an LLM rather than regular expressions, which can improve compatibility across future \greenagents that describe similar tasks using different instruction styles.

\subsection{Developing \GreenAgents}

For assessment designers, the first decision is whether to adopt a deterministic workflow to drive evaluation, sometimes even without an LLM, or to incorporate an agentic procedure where parts of the workflow are specified through prompts. For example, incorporating evaluations such as terminal-bench~\cite{merrill2025terminalbench} may benefit from a workflow-style design since evaluation is driven by scripts, while incorporating evaluations such as tau-bench~\cite{yao2024tau} may benefit more from prompting because user simulation is required. The primary challenge is to design the interaction procedure during assessment execution.

Following the interaction pattern in Section \ref{sec:aaa}, we recommend organizing the design process around the following questions:

\begin{itemize}[leftmargin=*, noitemsep, topsep=2pt]
    \item How is test-related data obtained and sampled?
    \item How is the evaluation environment prepared and initialized, and how is data loaded?
    \item How should the task be described, and how should environment be accessed? How should MCP tools be designed and exposed?
    \item How should the \greenagent interact with \whiteagents throughout assessment?
    \item During and after task fulfillment, what metrics should be used to score \whiteagent performance?
\end{itemize}

Careful consideration of these aspects improves compatibility, supports a wider range of \whiteagents, and provides a smoother assessment experience. For example:

\begin{itemize}[leftmargin=*, noitemsep, topsep=2pt]
    \item It can be useful to make dataset selection configurable through the kick-off request format, so that the requester can control expected runtime and scope.
    \item Avoiding repeated environment rebuilds by supporting reset semantics, and reducing unnecessary data transmission, can improve evaluation efficiency.
    \item Choosing appropriate tasks and scoring granularity can help \whiteagents better understand objectives, while clear tool descriptions help improve performance.
    \item Using a sub-agent inside the \greenagent can help isolate scoring logic from simulated user conversations, reducing the risk of prompt-injection-based cheating.
    \item Including metadata such as sampled data point identifiers and dataset size enables aggregation across runs.
\end{itemize}

Note that beyond building \greenagents for new scenarios and datasets, the process of agentifying existing benchmarks follows the same design procedure. The key challenge is to map the benchmark’s data and evaluation logic onto the corresponding components discussed above.

\section{Extension and Integrations}
\label{sec:ext}

The AAA paradigm also makes it easier to introduce additional abstractions that improve agent development and evaluation utilities. In this section, we discuss potential extensions to the AgentBeats implementation, as well as possible integrations. Some extensions are necessary or particularly useful for enabling features in certain operation modes. Others focus on improving openness and collaboration, or emphasize observability, providing greater transparency and assessment insights that support agent development and iteration.

\textbf{Agent Control Plane.} Agent blueprints in hosted mode provide transparency at the cost of privacy and deployment overhead, while agent instances in remote mode preserve privacy at the cost of reduced visibility. For example, when a remote agent instance exhibits errors or poor performance, execution logs are often unavailable. Similarly, when switching to a new assessment, there is typically no standardized mechanism to reset a remote agent service. To bridge this gap, AgentBeats implements an agent control plane that serves both as an operation extension and an observability layer. It is implemented as a thin service wrapped around the A2A service, providing APIs that allow a centralized platform to restart the agent process and stream execution logs back to the platform.

\textbf{Agent and Tool Authentication.} Concerns about abuse of publicly exposed agent services can be addressed through additional authentication components for agents and tools. Such functionality is essential for building a practical, production-ready platform that supports controlled access.

\textbf{MCP Gateway.} To address the dynamic MCP configuration challenge discussed earlier, one solution is to introduce an MCP gateway service with a fixed endpoint that supports dynamic control over tool access. As an operation extension, a \greenagent can connect its dynamically created MCP server as a provider to this gateway and selectively expose tools to authenticated \whiteagents. In this design, as long as a \whiteagent connects to the gateway using credentials tied to its agent identifier, dynamic MCP reconfiguration is handled transparently and no longer needs to be implemented inside the \whiteagent.

\textbf{Sandbox and Container Manager.} A shared sandbox and container management service can be introduced to assist \greenagents. This service centrally manages computational resources and avoids complex docker-in-docker setups when \greenagents are themselves containerized. Additionally, the manager may expose an MCP interface, making it easier to design \greenagents using prompt-driven workflows that require dynamic environment control.

\textbf{Agent Registry.} An agent registry is a collaboration-focused extension that supports discovery of both \whiteagents and benchmarks that are represented as \greenagents. It maintains a publicly accessible list of platform identifiers and associated metadata, potentially derived from A2A agent cards, enabling users to efficiently select appropriate \greenagents and \whiteagents for a given assessment.

\textbf{Agent Leaderboard.} To provide clearer visibility into agent performance and support ranking among multiple \whiteagents evaluated under the same \greenagent, an agent leaderboard can be introduced. To improve interoperability, the leaderboard may also define recommended metric reporting formats, simplifying \greenagent response parsing and result aggregation.

Beyond extending AgentBeats directly, the AAA paradigm is also integration-friendly and can interoperate with a range of third-party services to improve overall utility. For example, it can integrate with \citeauthor{huggingface} for richer leaderboard functionality, \citeauthor{litellm} and \citeauthor{mlflow} for improved LLM access observability, and \citeauthor{wandb} for logging and experiment management. 

\textbf{Reinforcement Learning.} AAA is also well suited for reinforcement learning (RL) integration. 
To bridge the gap between Gymnasium-style APIs commonly used in RL and the protocol-based interaction model of AAA, AgentBeats supports translation from type-specified Gym interfaces to self-descriptive MCP tools, enabling any MCP-compatible agent to explore RL environments without native Gym support. 
Combined with environment-management systems such as OpenEnv~\cite{openenv} and existing RL training infrastructure, AAA provides a universal interaction standard, while also introducing an intermediate layer that facilitates tracking and analysis of agent behavior across training and evaluation.

\textbf{Agent Governance.} AgentBeats can also be combined with emerging work on agent governance and policy enforcement. Beyond measuring task success alone, evaluations may incorporate governance constraints such as audit requirements, resource budgets, restricted actions, permission policies, disclosure obligations, sanctions, or human-approval gates. Such settings enable assessments of how agents behave under different governance regimes, rather than only whether they complete a task. Because AgentBeats standardizes task execution and evaluation workflows, it can provide a reproducible substrate for studying the interaction between agent capabilities and governance mechanisms. The resulting traces may further support experimental analyses of how different policy settings influence agent behavior and performance, turning agent assessments into reproducible experiments on agentic governance.

\textbf{CUBE.} Concurrent with AgentBeats, CUBE~\cite{lacoste2026cubestandardunifyingagent} is an independent standardization effort targeting the benchmark \emph{infrastructure packaging} layer. While AgentBeats standardizes how evaluation campaigns are orchestrated, namely which agents evaluate which subjects, through which protocols, and under which deployment mode, CUBE standardizes how individual benchmark environments are packaged, deployed, reset, and lifecycle-managed, independently of any evaluation framework. Concretely, CUBE defines a protocol through which a benchmark exposes a task-spawning interface (\texttt{cube/spawn}), a Gym-style interaction API (\texttt{cube/reset}, \texttt{cube/step}, \texttt{cube/evaluate}), and infrastructure lifecycle primitives over an MCP-compatible JSON-RPC transport. The two standards are naturally composable: a \greenagent can use a CUBE-compliant benchmark as its internal infrastructure substrate, calling \texttt{cube/spawn} to instantiate task environments per \whiteagent, distributing the resulting MCP tool endpoints via A2A, and invoking \texttt{cube/evaluate} to obtain scores. Conversely, the CUBE registry of packaged benchmarks provides a natural catalog of candidates for agentification as \greenagents.

\section{Field Study on Real-World Adoption}
\label{sec:adoption}

\begin{figure*}[t]
    \centering
    \includegraphics[width=0.95\textwidth]{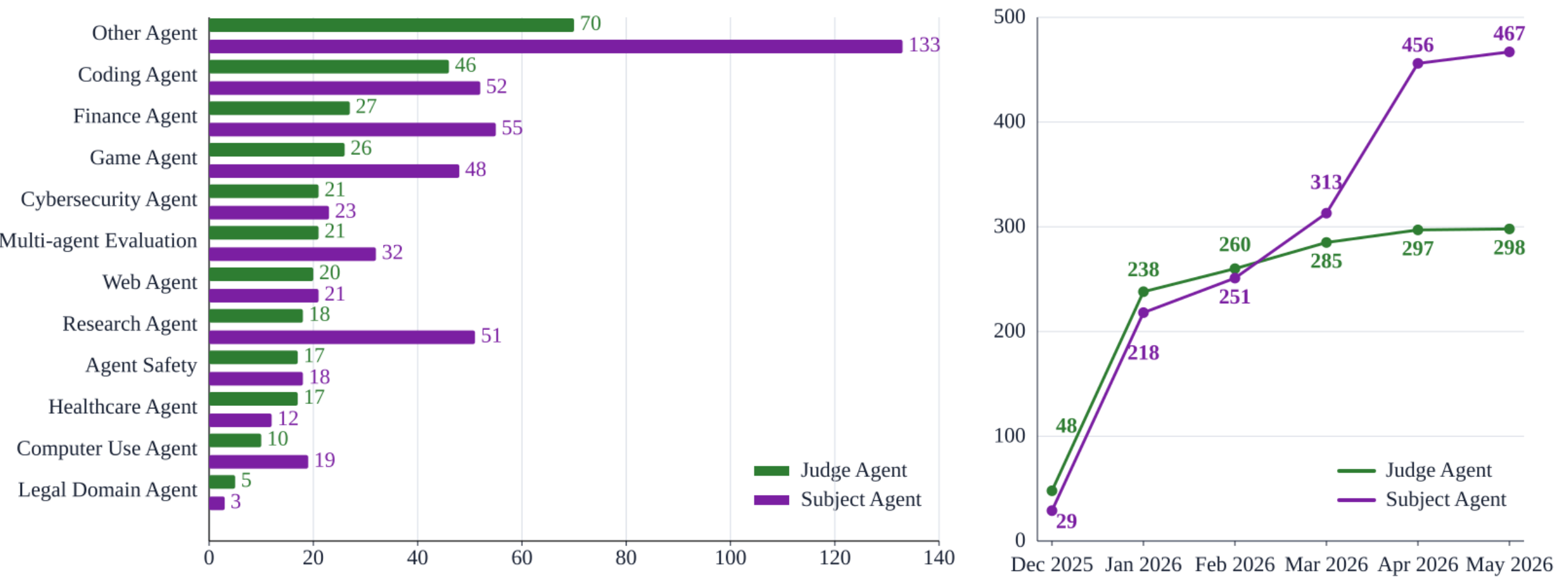}
    \caption{Competition submissions: distribution of \greenagents/\whiteagents across categories (left) and submission timeline over the course of the competition (right).}
    \label{fig:adoption}
\end{figure*}

Having presented the AAA paradigm, AgentBeats, and its extensions, we now evaluate the approach at scale. The most direct way to assess whether the AAA paradigm is workable is to observe how real agent developers use it in practice. We wrote documentation for the AAA conventions, built the AgentBeats platform (Section~\ref{sec:ab}) for submitting and evaluating AAA-compatible agents in CI mode, and ran the AgentX-AgentBeats competition\footnote{\url{https://rdi.berkeley.edu/agentx-agentbeats.html}}, held in conjunction with the Agentic AI MOOC and its global community of around 40,000 registered learners. The resulting field study covers 298 \greenagents across 12 categories, together with 467 \whiteagents from independent participants. The remainder of this section describes the platform and competition setup, summarizes the agent statistics, and reports insights from the submissions.

\paragraph{Competition preparation, submission platform, and logistics.}
To make the paradigm easy to adopt, we wrote documentation describing the AAA conventions in detail, and provided an official example based on tau-bench~\cite{yao2024tau}, accompanied by a blog series that walks through its end-to-end conversion as a reference for community contributors. To support the competition and collect statistics, the platform accepts repository-level submissions of A2A-compatible agents through dedicated assessment repositories. Each assessment repository links to the agents under test, selects which subjects to run, and drives their interaction through GitHub Actions. The platform also tracks results on a shared leaderboard maintained by the corresponding \greenagent, so that \whiteagents can be cross-compared. To encourage adoption, we offered prizes for top submissions, inviting developers to build \greenagents (either for new benchmarks of their own design, or by agentifying existing ones) and to register \whiteagents as opponents. All submitted agents were required to follow the AAA paradigm and conform to the A2A and MCP standards. The competition is structured into multiple phases over six months; at the time of writing, five months have elapsed.
\paragraph{Submission analysis.}
Figure~\ref{fig:adoption} (left) shows the distribution of the 298 \greenagents across the 12 competition categories, which span diverse agent capabilities. So far, the competition has run two phases: Phase~1 (October through late January) focused on building \greenagents, while Phase~2 (from March) invited more \whiteagents; this shift is visible in the timeline on the right.

Within these submissions, many well-known existing benchmarks were agentified, including Tau2-Bench~\cite{barres2025tau2benchevaluatingconversationalagents}, MedAgentBench~\cite{jiang2025medagentbenchrealisticvirtualehr}, FinanceAgent~\cite{bigeard2025financeagentbenchmarkbenchmarking}, OfficeQA~\cite{opsahlong2026officeqaproenterprisebenchmark}, and PersonaGym~\cite{samuel2025personagymevaluatingpersonaagents}, among others. This breadth, spanning tool use, coding, browser interaction, multi-agent games, healthcare, and cybersecurity, indicates that the AAA interface is expressive enough to accommodate heterogeneous evaluation scenarios without requiring fundamental redesign of existing benchmark logic. The high uptake of multi-agent evaluation in particular suggests that AAA naturally supports coordination-intensive assessments that are traditionally awkward to set up. The successful agentification of benchmarks originally built around tightly coupled harnesses (e.g., OSWorld~\cite{xie2024osworld}) further suggests that the migration overhead, while real, is manageable in practice.

Beyond the \greenagents, the platform also hosts 467 \whiteagents from independent developers, with finance, research, coding, and game agents being the most represented categories. The platform's submission flow is light enough that developers iterate on it routinely. Tau2-Bench, for instance, received 347 assessments from 42 unique \whiteagents, with 16 developers each submitting 10 or more versions of their agent.

\begin{table}[h]
  \centering
  \caption{Python code and prompt characteristics of submitted agents. ``\%~repos'' denotes the portion of repos that include any prompts, while ``chars'' shows the prompt character counts in those repos.}
  \label{tab:judge-impl}
  \small
  \resizebox{\columnwidth}{!}{%
  \begin{tabular}{lrrrrr}
    \toprule
                     & \multicolumn{1}{c}{Python} & \multicolumn{4}{c}{Prompts} \\
    \cmidrule(lr){2-2} \cmidrule(lr){3-6}
    Agent role       & Avg LOC & \%~repos & Avg LOC & Avg chars & Avg chars\% \\
    \midrule
    \Greenagents & 5.3k & 78.3\% &  241 & 11.6k & 5.00\% \\
    \Whiteagents & 3.8k & 87.1\% &  184 &  8.7k & 7.35\% \\
    Total & 4.1k & 81.7\% &  192 &  9.3k & 6.66\% \\
    \bottomrule
  \end{tabular}%
  }
\end{table}

\paragraph{Implementation analysis.}
Diving into the submitted code, we collected the repositories, excluded inaccessible or malformed ones, and deduplicated the rest. Table~\ref{tab:judge-impl} reports the analysis on the lines of code\footnote{We manually filter out files unrelated to agent logic, such as JSON/CSV for data, YAML for configuration, etc.}. Python remains the most popular language (99\% in \greenagents, 98\% in \whiteagents), but we also find agents purely implemented in other languages, including TypeScript and Rust, a flexibility afforded by AAA's interface, which constrains the protocol rather than the language.

As discussed in Section~\ref{sec:aaa}, semantic internalization allows a \greenagent to declare its evaluation logic through prompts. To see how prevalent this is across submissions, we built a prompt-extracting code-scanning tool and report its estimates in the same table. About 78\% of \greenagents contain at least one extracted prompt, and among those, prompts average 241 LOC (11.6k characters), roughly 5\% of code by character count. We also observe a heavy-prompt tail, with some agents devoting nearly a third of their code characters to prompts, pointing to a meaningful subpopulation of \greenagents built primarily through natural-language specification. Overall, prompts remain a common component of \greenagents, suggesting that explicit support for semantic declaration of evaluation logic is worth considering for future benchmark-standardization effort.

\section{Zoom-in Case Study on Coding Agents}
\label{sec:casestudy}

Our field study demonstrated the coverage and practicality of AAA, leaving two questions: whether agentified evaluation preserves fidelity, and whether it can yield meaningful research insights. In this section we examine both through a concrete example, evaluating coding agents under a single AAA-standardized pipeline. We find that the standardized pipeline preserves fidelity and surfaces research insights, including head-to-head comparisons the public record currently cannot deliver, while also concretely instantiating our earlier recommended practices. More case studies on agentifying existing benchmarks can be found in Appendix~\ref{app:benchmark-case-studies}.

\subsection{Agentifying Coding-Agent Benchmarks}

Coding-agent benchmarks exercise a part of AAA that earlier sections only touched on. Among our recommended practices (Section~\ref{sec:rec}), we noted that some assessments require capabilities beyond A2A and MCP, such as a prepared development environment or remote machine access; coding-agent benchmarks make this concrete. Such benchmarks require the \whiteagent to operate inside a specific environment that must be set up before the agent runs: the agent needs to work inside a particular development environment and repository checkout, so the \whiteagent should be launched only after that environment is in place. In a traditional workflow this is handled implicitly, because the benchmark harness launches and manages the agent itself. Under AAA, however, the \whiteagent is launched independently and decoupled from any specific benchmark; this decoupling improves interoperability but complicates post-launch environment configuration, since the \whiteagent must accept an environment provided after it starts, which not every agent natively supports.

We adopt a design pattern that divides responsibilities so that the \greenagent prepares and controls the environment while the \whiteagent only needs to receive the task and respond when it is done. The \greenagent first sets up an isolated container with the source tree, dependencies, and evaluation scripts in place. By default, the \whiteagent then copies itself, along with its runtime, credentials, and configuration, into that container and runs there as if it were its own environment, so that file-system and terminal access match how the agent is normally used. Because this injection step can fail or perturb the environment, we guard it with a preflight check. When a \whiteagent cannot copy itself in, the \greenagent instead exposes the same container as a remote-shell tool over MCP, and the \whiteagent drives it from outside. In either mode, once the \whiteagent signals completion, the \greenagent collects the artifacts, runs the benchmark's evaluation script, and returns the final score. This pattern accommodates \whiteagents that need a runtime environment supplied at evaluation time while keeping them decoupled from any specific benchmark.

Our case study evaluates four agents, each pairing a model with an agent harness. Three pair proprietary models with production-ready harnesses: Claude Opus 4.7 $\times$ Claude Code~\cite{anthropic2026opus47, anthropic2026claudecode}, GPT-5.4 $\times$ Codex CLI~\cite{openai2026gpt54, openai2026codex}, and Gemini 3.1 Pro $\times$ OpenCode~\cite{googledeepmind2026gemini31pro, opencode2026}. The fourth, Qwen3.5-397B-A17B $\times$ mini-SWE-agent (MSA)~\cite{qwen3.5, yang2024sweagent}, pairs an open-weight model with a research-oriented harness as our open-source baseline. Opus 4.7 and GPT-5.4 use their providers' native harnesses, while Gemini 3.1 Pro is among the recommended models for OpenCode. We abbreviate the models as Opus 4.7, GPT-5.4, Gemini 3.1 Pro, and Qwen3.5 throughout. We evaluate these agents on three coding benchmarks: DevEval~\cite{li2024deveval} for function-level completion, SWE-Bench Pro~\cite{deng2025swebenchpro} for repository-level issue resolution, and Terminal-Bench 2.0~\cite{merrill2025terminalbench} for terminal-mediated workflows.

\subsection{Incomplete Public Records}
\label{subsec:missing-eval}

Despite the rapid proliferation of the aforementioned coding agents, their public performance records are scattered across vendor reports and leaderboards, often incomplete, and hard to compare directly. 
We conduct a best-effort survey of first-party publicly accessible assessment records. Table~\ref{tab:public-reference-scores} compiles all comparable scores.

\begin{table}[h]
  \centering
  \caption{Public records of coding agent performance on SWE-Bench Pro and Terminal-Bench 2.0. No DevEval scores are available for the selected agents. Proprietary harnesses are not publicly accessible.}\label{tab:public-reference-scores}
  \small
  \resizebox{\columnwidth}{!}{%
  \begin{tabular}{lllll}
    \toprule
    Benchmark & Model & Harness & Public score & Source \\
    \midrule
    SWE-Bench Pro & Claude Opus 4.7 & Proprietary & 64.3\% & \cite{anthropic2026opus47} \\
                  & Claude Opus 4.6 & MSA & 51.9\% & \cite{swebenchpro_leaderboard} \\
                  & GPT-5.4 & MSA & 59.1\% & \cite{swebenchpro_leaderboard} \\
                  & Gemini 3.1 Pro & Proprietary & 54.2\% & \cite{googledeepmind2026gemini31pro} \\
                  & Gemini 3.1 Pro & MSA & 46.1\% & \cite{swebenchpro_leaderboard} \\
                  & Qwen3.5-397B-A17B & Proprietary & 50.9\% & \cite{qwen36plus}\\
    \addlinespace[0.35em]
    Terminal-Bench 2.0 & Claude Opus 4.7 & Terminus-2 & 69.4\% & \cite{anthropic2026opus47} \\
                   & Claude Opus 4.6 & Claude Code & 58.0\% & \cite{merrill2025terminalbench} \\
                   & Claude Opus 4.6 & OpenCode & 52.1\% & \cite{merrill2025terminalbench} \\
                   & GPT-5.4 & Proprietary & 75.1\% & \cite{openai2026gpt54} \\
                   & GPT-5.2 & Codex CLI & 62.9\% & \cite{merrill2025terminalbench} \\
                   & Gemini 3.1 Pro & Terminus-2 & 68.5\% & \cite{googledeepmind2026gemini31pro} \\
                   & Qwen3.5-397B-A17B & Terminus-2 & 52.5\% & \cite{qwen36plus} \\
    \bottomrule
  \end{tabular}%
  }
\end{table}

We highlight three issues. First, scores for the same model can shift substantially when the agent and benchmark sides use different harnesses: on SWE-Bench Pro, Gemini's official score under a proprietary harness is 8 percentage points above the benchmark team's score for the same model under MSA, and more similar gaps can be found in the table. Second, proprietary harnesses make these rankings hard to verify: the Opus 4.7, Gemini 3.1 Pro, and Qwen3.5 reports for SWE-Bench Pro all rely on proprietary harnesses that no third party can rerun, and reproduced numbers can diverge sharply, with a separate team~\cite{moonshot2026kimik26} reporting 65.4\% for GPT-5.4 under Terminus-2, almost 10 points below the vendor-reported number. Third, public scores for production-ready harnesses are scarce, usually limited to a vendor's in-house pairing, with many model versions still missing.

Together, these gaps make fair, side-by-side comparison difficult. To address them, we leverage the AAA paradigm (Section~\ref{sec:aaa}) to standardize agent benchmarking, evaluating the model-harness pairs under settings that are fully disclosed and match what users actually run in practice.

\subsection{Experiment Preparation}

For SWE-Bench Pro, we evaluate on the 731 instances of the public split; for Terminal-Bench 2.0, we use all instances. We adapt DevEval into an agentic setting by building a containerized development sandbox analogous to those in SWE-Bench Pro: each sandbox is seeded with the source codebase, with the test suite hidden. We then verify ground-truth solutions by running the test suites in our environment and drop any failing instances, yielding 1{,}222 DevEval instances. Because the benchmarks are fully decoupled from the agent harness, we author meta instructions that brief each agent on the evaluation workflow. The \greenagent provisions the \whiteagent's environment using the design pattern described above.

Each agent runs with its default prompt, reasoning setting, and inference configuration, matching typical deployment use. The thinking level is set to ``xHigh'' for Opus 4.7, ``Medium'' for GPT-5.4, and ``High'' for Gemini 3.1 Pro, while Qwen3.5 runs with thinking enabled and no further effort setting. The full experiment cost around \$6{,}000.

\subsection{Assessing Representative Coding Agents on Agentified Benchmarks}
\label{subsec:comparison}

\begin{table}[t]
  \centering
  \caption{Performance of representative coding agents on three different agentified benchmarks. Qwen3.5 model is self-hosted so it has no associated API cost.}
  \label{tab:main-results}
  \small
  \resizebox{\columnwidth}{!}{%
  \begin{tabular}{lllrr}
    \toprule
    Benchmark & Model & Harness & Solve Rate & Avg Cost/inst. \\
    \midrule
    DevEval & Claude Opus 4.7 & Claude Code & 91.9\% & \$0.26 \\
            & GPT-5.4 & Codex & \textbf{94.8\%} & \$0.20 \\
            & Gemini 3.1 Pro & OpenCode & 88.8\% & \$0.52 \\
            & Qwen3.5-397B-A17B & MSA & 71.4\% & -- \\
    \addlinespace[0.35em]
    SWE-Bench Pro & Claude Opus 4.7 & Claude Code & \textbf{69.1\%} & \$2.28 \\
                  & GPT-5.4 & Codex & 62.1\% & \$1.41 \\
                  & Gemini 3.1 Pro & OpenCode & 54.4\% & \$0.97 \\
                  & Qwen3.5-397B-A17B & MSA & 45.8\% & -- \\
    \addlinespace[0.35em]
    Terminal-Bench 2.0 & Claude Opus 4.7 & Claude Code & \textbf{68.5\%} & \$1.09 \\
                   & GPT-5.4 & Codex & 67.4\% & \$0.64 \\
                   & Gemini 3.1 Pro & OpenCode & 65.2\% & \$0.71 \\
                   & Qwen3.5-397B-A17B & MSA & 36.0\% & -- \\
    \bottomrule
  \end{tabular}%
  }
\end{table}

\paragraph{No single agent system leads across all three benchmarks.}
GPT-5.4 $\times$ Codex CLI achieves the best result on DevEval, while Opus 4.7 $\times$ Claude Code leads on SWE-Bench Pro and Terminal-Bench 2.0; on Terminal-Bench 2.0, the two are nearly tied. The benchmarks also differ in distinguishing power: the top proprietary systems are tightly clustered on Terminal-Bench 2.0, spanning only 3.3 percentage points, whereas SWE-Bench Pro separates them more clearly with successive gaps of about 7 percentage points. The open-weight Qwen3.5 $\times$ MSA achieves non-trivial solve rates across all benchmarks but remains consistently behind the leaders.

\paragraph{Per-instance costs cluster within an order of magnitude.}
Across all four agents and three benchmarks, the most and least expensive systems differ by at most 2.4$\times$ per instance, well within an order of magnitude. Opus 4.7 $\times$ Claude Code, the strongest performer on SWE-Bench Pro and Terminal-Bench 2.0, is also the most expensive proprietary system on those two benchmarks. Across all agents, SWE-Bench Pro and Terminal-Bench 2.0 are systematically more expensive than DevEval. Gemini 3.1 Pro $\times$ OpenCode exhibits a striking cost anomaly on DevEval: it has the highest DevEval cost among proprietary systems despite achieving the lowest proprietary solve rate. Trajectory inspection points to inefficient repository exploration and model-harness incompatibility, with a tool-call error rate of 2\%, more than 10 times higher than the other agents.

\paragraph{Gap between ours and public scores.}
Comparing our results with the public records in Table~\ref{tab:public-reference-scores}, many of our scores closely match the public record, but a few show notable drifts. In our results, the model ranking on both SWE-Bench Pro and Terminal-Bench 2.0 is Opus 4.7 $>$ GPT-5.4 $>$ Gemini 3.1 Pro $>$ Qwen3.5. This ordering matches the public SWE-Bench Pro reports. The main exception is Terminal-Bench 2.0, where the public report places GPT-5.4 ahead of Opus 4.7 by roughly 6 percentage points under a proprietary harness; under our consistent model-harness setting, Opus is ahead by 1.1 points, suggesting that the proprietary harness drives most of the public lead. Three sources of mismatch in evaluation settings explain the largest absolute-score drifts. Anthropic's reported Opus 4.7 score on SWE-Bench Pro applies a memorization screen that flags a subset of problems, which we do not replicate~\cite{anthropic2026mythos}. Qwen reports its score on a corrected version of SWE-Bench Pro that is not publicly released in the same form~\cite{qwen36plus}. For GPT-5.4 on Terminal-Bench 2.0, the official report uses a higher thinking effort (``xhigh''), while we use the harness default (``medium''). Our goal is not to reproduce every public number exactly, but to provide an internally consistent head-to-head comparison under transparent evaluation conditions.

\subsection{Harness-Swapping Experiment}

Frontier coding models are reportedly optimized for use with their native harness~\cite{openai2025gpt51codexmax}, but there is no consistent measurement of how well their proficiency generalizes across harness choices. We conduct a harness-swapping experiment by pairing GPT-5.4 with Claude Code and Opus 4.7 with Codex.\footnote{To bridge the API differences between the two stacks, we route requests through a gateway layer for translation.}

\begin{table}[t]
  \centering
  \caption{Harness-swapping experiment. A dagger marks each model's native harness. To control cost, the DevEval and SWE-Bench Pro runs use a 20\% subsample of the full main experiment, while Terminal-Bench 2.0 uses all 89 instances, given its small total.}
  \label{tab:model-harness-ablations}
  \small
  \resizebox{\columnwidth}{!}{%
  \begin{tabular}{lllrrr}
    \toprule
    Benchmark & Model & Harness & Solve rate & Input tokens & Output tokens \\
    \midrule
    DevEval & Claude Opus 4.7 & Claude Code$^\dagger$ & 93.9\% & 30.652M & 1.505M \\
            &                   & Codex & 87.9\% & 42.798M & 0.560M \\
            & GPT-5.4 & Codex$^\dagger$ & \textbf{96.0\%} & 106.681M & 2.139M \\
            &         & Claude Code & 88.7\% & 41.319M & 2.601M \\
    \addlinespace[0.35em]
    SWE-Bench Pro & Claude Opus 4.7 & Claude Code$^\dagger$ & \textbf{71.9\%} & 332.599M & 4.371M \\
                  &                   & Codex & 67.8\% & 158.021M & 1.207M \\
                  & GPT-5.4 & Codex$^\dagger$ & 61.0\% & 383.759M & 3.360M \\
                  &         & Claude Code & 48.6\% & 90.395M & 2.616M \\
    \addlinespace[0.35em]
    Terminal-Bench 2.0 & Claude Opus 4.7 & Claude Code$^\dagger$ & 68.5\% & 42.878M & 1.912M \\
                   &                   & Codex & \textbf{71.9\%} & 60.280M & 0.984M \\
                   & GPT-5.4 & Codex$^\dagger$ & 67.4\% & 100.907M & 1.382M \\
                   &         & Claude Code & 61.8\% & 16.140M & 0.776M \\
    \bottomrule
  \end{tabular}%
  }
\end{table}

\paragraph{Co-adaptation verified in solve rate, with possible trade-offs in solve speed.}
Table~\ref{tab:model-harness-ablations} shows that native pairings perform best in five of the six comparisons, with the original set outperforming the swapped one by an average of 5.3 percentage points. The one exception is Opus 4.7 on Terminal-Bench 2.0. A closer look attributes this to solve speed: within the standard time budget Codex finishes more tasks, while given sufficient time Opus 4.7 $\times$ Claude Code reaches 79.8\% and Opus 4.7 $\times$ Codex reaches 73.6\%, restoring the native advantage. Native pairings therefore win on solve rate when solve speed is not a bottleneck, while the swapped harness can potentially be faster.

\paragraph{GPT-5.4 uses more tokens under its native Codex harness.}
Per-instance cost is difficult to compare directly across the swapped pairings, since pricing and prompt caching differ between providers. We therefore compare token usage, focusing on input tokens, which dominate total spend in our experiments. Within this view, GPT-5.4's native pairing (Codex CLI) consumes more input tokens per instance than its swapped pairing (Claude Code) on every benchmark we measure. Trajectory inspection suggests this might be driven by Codex reading larger chunks of files at a time.

\section{Conclusion}

Agent evaluation today suffers from severe compatibility issues due to the lack of standardization, while limited openness and reproducibility further hinder progress. Agentifying the agent assessment is a necessary step forward, and provides a practical roadmap to fundamentally improve agent benchmarking practice. Our community-scale field study, with hundreds of independent developers, demonstrates the coverage and practicality of the AAA paradigm and AgentBeats, while our standardization case study on coding agents demonstrates fidelity and surfaces concrete insights about representative coding agents. 
Agentifying the agent assessment is a necessary step forward, and provides a practical roadmap to fundamentally improve agent benchmarking practice.

\section{Frequently Asked Questions}

\begin{itemize}
    \item \textbf{Does AgentBeats require agents to support dynamic loading of new MCP servers?}
No. AgentBeats is designed to work with general-purpose agents and does not assume any special capability such as runtime MCP reconfiguration. To address the dynamic tool-access challenge, AgentBeats incorporates an MCP gateway with a fixed endpoint that supports dynamic registration and de-registration of tools (Section~\ref{sec:ext}). A \whiteagent connects to the gateway using a static URL during setup. During evaluation, the \greenagent can register task-specific tools on demand and selectively grant access to the \whiteagent through the gateway, enabling flexible tool provisioning without requiring any changes to the \whiteagent.
    \item \textbf{How can terminal-focused agents, such as coding agents, be evaluated under AAA?}
As discussed in Section~\ref{sec:rec}, a common pattern is to build a thin A2A wrapper that forwards task instructions to the agent's terminal interface, captures its output, and returns the result as an A2A response. When a \greenagent provides a remote environment accessible via SSH, \whiteagents can either implement tools that translate A2A interactions into remote SSH command execution, or transmit the agent executable directly to the target environment and monitor its execution remotely.
    \item \textbf{How can browser-use and computer-use agents be evaluated under AAA?}
Evaluation of browser-use and computer-use agents can generally be categorized into two types: goal-oriented and procedure-oriented. In goal-oriented assessments, the \greenagent does not require explicit access to the browser or computer environment, but instead focuses solely on whether the target task, typically one that inherently involves browser or computer use, has been successfully completed. In such cases, the \greenagent provides clear task instructions along with any necessary attachments, while \whiteagents are responsible for implementing the required mechanisms to interact with the environment.
In contrast, procedure-oriented assessments place greater emphasis on the execution process itself. The \greenagent may provide its own execution environment or define evaluation metrics that depend on intermediate steps. In these settings, the \greenagent should expose MCP tools that offer explicit access to browser or computer states and actions, potentially across multiple modalities. Alternatively, such procedural information can also be conveyed through continuous A2A interactions. We provide details for a few concrete cases in Appendix~\ref{app:benchmark-case-studies}.
    \item \textbf{Does the AAA interface affect model performance on specific tasks?}
A common concern is that a general-purpose interface introduces overhead that may degrade task-specific performance. While the interface does add a thin processing layer, AgentBeats does not prevent task-specific optimizations. For example, a \whiteagent developer can detect certain task instructions within an A2A message and trigger specialized execution branches that invoke task-specific logic, achieving the same performance as running directly against the original benchmark harness. In this sense, the interface design itself does not bottleneck performance improvement. Findings from Section~\ref{subsec:comparison} also verify this. At the same time, AgentBeats targets general-purpose agents that can compete across multiple benchmarks under a unified interface, thereby expanding the generalizability and coverage of each benchmark.
    \item \textbf{What is the computational overhead of evaluation under AAA?}
By default, AAA adds only a very thin layer on top of the underlying evaluation. When a \greenagent uses a fully deterministic program and connects directly to the \whiteagent, the only additional cost is wrapping interactions as A2A and MCP messages. This wrapping is on the order of milliseconds per message, which is negligible compared to the seconds-scale compute and latency of LLM inference. However, because we want a \greenagent to be compatible with as many \whiteagents as possible, the \greenagent itself often needs to be intelligent, which can introduce extra LLM inference cost. The interoperability-versus-cost trade-off is therefore an important consideration when designing a \greenagent. In the common case where most LLM inference is spent on the task itself rather than on orchestration, this is a favorable trade-off, though the precise balance is benchmark-specific and should be analyzed case by case. As a concrete reference point, Appendix~\ref{app:additional_results} reports per-instance tool-call counts from our coding-agent experiments; a natural future step is to compare these against other public records.

\item \textbf{As an agent developer, what are the benefits of adopting AgentBeats?}

AgentBeats allows developers to evaluate the same agents they build and deploy, rather than maintaining benchmark-specific integrations or adapted evaluation versions. Once an agent supports standard interfaces such as A2A and MCP, it can participate in a broad range of compatible assessments with minimal additional engineering effort. This lowers the cost of benchmarking, makes it easier for third parties to evaluate and compare agents, and enables results to transfer more directly to real-world deployments.
\url{https://agentbeats.org} can serve as a default platform for agent registration and evaluation, but the AAA paradigm itself is deployment-agnostic and can be integrated into existing internal evaluation pipelines. For organizations that already expose A2A-compatible agents, adoption is typically straightforward because assessments interact with the same interfaces used in production. 

\end{itemize}

\clearpage


\bibliographystyle{ACM-Reference-Format}
\bibliography{content/ref}

@misc{nemoevaluator,
  title        = {NeMo Evaluator},
  author       = {{NVIDIA}},
  howpublished = {https://github.com/NVIDIA-NeMo/Evaluator},
  year         = {2026}
}

@misc{openhandsbenchmarks,
  title        = {OpenHands Repository},
  author       = {{OpenHands}},
  howpublished = {https://github.com/OpenHands/OpenHands/tree/main/evaluation/benchmarks},
  year         = {2026}
}

@misc{a2a2025specification2025,
  title        = {Agent2Agent (A2A) Protocol Specification},
  author       = {A2A},
  howpublished = {Online draft specification},
  year         = {2025},
  note         = {Draft v0.3.0},
  url          = {https://a2a-protocol.org/latest/specification/},
  organization = {A2A Protocol Project}
}

@misc{mcp2025specification,
  title        = {Model Context Protocol (MCP) Specification},
  author       = {MCP},
  howpublished = {Online technical specification},
  year         = {2025},
  url          = {https://modelcontextprotocol.io/specification/2025-11-25},
  organization = {Model Context Protocol Project}
}

@misc{kaggle2025gamearena,
  title        = {Kaggle Game Arena},
  author       = {Kaggle},
  year         = {2025},
  url          = {https://www.kaggle.com/game-arena},
  note         = {A benchmarking platform where AI models and agents compete in strategic games},
  organization = {Kaggle}
}

@article{xu2023language,
  title   = {Language agents with reinforcement learning for strategic play in the werewolf game},
  author  = {Xu, Zelai and Yu, Chao and Fang, Fei and Wang, Yu and Wu, Yi},
  journal = {arXiv preprint arXiv:2310.18940},
  year    = {2023}
}

@article{zou2026ctfagent,
  title     = {CTFAgent: An LLM-powered Agent for CTF Challenge Solving},
  author    = {Zou, Yuwen and Liu, Jia and Fan, Wenjun},
  journal   = {Journal of Information Security and Applications},
  volume    = {96},
  pages     = {104305},
  year      = {2026},
  publisher = {Elsevier}
}

@article{wei2022chain,
  title   = {Chain-of-thought prompting elicits reasoning in large language models},
  author  = {Wei, Jason and Wang, Xuezhi and Schuurmans, Dale and Bosma, Maarten and Xia, Fei and Chi, Ed and Le, Quoc V and Zhou, Denny and others},
  journal = {Advances in neural information processing systems},
  volume  = {35},
  pages   = {24824--24837},
  year    = {2022}
}

@inproceedings{yao2022react,
  title     = {React: Synergizing reasoning and acting in language models},
  author    = {Yao, Shunyu and Zhao, Jeffrey and Yu, Dian and Du, Nan and Shafran, Izhak and Narasimhan, Karthik R and Cao, Yuan},
  booktitle = {The eleventh international conference on learning representations},
  year      = {2022}
}

@article{yao2024tau,
  title   = {Tau-bench: A Benchmark for Tool-Agent-User Interaction in Real-World Domains},
  author  = {Yao, Shunyu and Shinn, Noah and Razavi, Pedram and Narasimhan, Karthik},
  journal = {arXiv preprint arXiv:2406.12045},
  year    = {2024}
}

@misc{harborframeworkteam2026harborframework,
  title        = {Harbor Framework: A framework for evaluating and optimizing agents and models in container environments.},
  author       = {{Harbor Framework Team}},
  year         = {2026},
  howpublished = {https://github.com/laude-institute/harbor}
}

@misc{hal,
  title        = {Holistic Agent Leaderboard: The Missing Infrastructure for AI Agent Evaluation},
  author       = {Sayash Kapoor and Benedikt Stroebl and Peter Kirgis and Nitya Nadgir and Zachary S Siegel and Boyi Wei and Tianci Xue and Ziru Chen and Felix Chen and Saiteja Utpala and Franck Ndzomga and Dheeraj Oruganty and Sophie Luskin and Kangheng Liu and Botao Yu and Amit Arora and Dongyoon Hahm and Harsh Trivedi and Huan Sun and Juyong Lee and Tengjun Jin and Yifan Mai and Yifei Zhou and Yuxuan Zhu and Rishi Bommasani and Daniel Kang and Dawn Song and Peter Henderson and Yu Su and Percy Liang and Arvind Narayanan},
  howpublished = {https://github.com/princeton-pli/hal-harness},
  year         = {2025}
}

@misc{huggingface,
  title        = {Hugging Face},
  author       = {{Hugging Face}},
  howpublished = {https://huggingface.co},
  note         = {Open platform for machine learning models, datasets, and evaluation},
  year         = {2026}
}

@misc{litellm,
  title        = {LiteLLM},
  author       = {{LiteLLM}},
  howpublished = {https://github.com/BerriAI/litellm},
  year         = {2026}
}

@misc{mlflow,
  title        = {MLflow},
  author       = {{MLflow}},
  howpublished = {https://github.com/mlflow/mlflow},
  year         = {2026}
}

@misc{wandb,
  title        = {Weights \& Biases},
  author       = {{Weights \& Biases}},
  howpublished = {https://wandb.ai},
  year         = {2026}
}

@misc{openenv,
  title        = {OpenEnv},
  author       = {{Meta}},
  howpublished = {https://github.com/meta-pytorch/OpenEnv},
  note         = {Open-source platform for environment-based agent training and evaluation},
  year         = {2026}
}

@misc{perplexitycomet,
  author = {{Perplexity AI}},
  title  = {Comet},
  url    = {https://www.perplexity.ai/comet},
  year   = {2026}
}

@misc{openaichatgptatlas,
  author = {OpenAI},
  title  = {ChatGPT Atlas},
  url    = {https://chatgpt.com/atlas},
  year   = {2026}
}

@article{sager2025comprehensive,
  title   = {A Comprehensive Survey of Agents for Computer Use: Foundations, Challenges, and Future Directions},
  author  = {Sager, Pascal J and Meyer, Benjamin and Yan, Peng and von Wartburg-Kottler, Rebekka and Etaiwi, Layan and Enayati, Aref and Nobel, Gabriel and Abdulkadir, Ahmed and Grewe, Benjamin F and Stadelmann, Thilo},
  journal = {arXiv preprint arXiv:2501.16150},
  year    = {2025}
}

@misc{barres2025tau2benchevaluatingconversationalagents,
  title         = {$\tau^2$-Bench: Evaluating Conversational Agents in a Dual-Control Environment},
  author        = {Victor Barres and Honghua Dong and Soham Ray and Xujie Si and Karthik Narasimhan},
  year          = {2025},
  eprint        = {2506.07982},
  archiveprefix = {arXiv},
  primaryclass  = {cs.AI},
  url           = {https://arxiv.org/abs/2506.07982}
}

@misc{jiang2025medagentbenchrealisticvirtualehr,
  title         = {MedAgentBench: A Realistic Virtual EHR Environment to Benchmark Medical LLM Agents},
  author        = {Yixing Jiang and Kameron C. Black and Gloria Geng and Danny Park and James Zou and Andrew Y. Ng and Jonathan H. Chen},
  year          = {2025},
  eprint        = {2501.14654},
  archiveprefix = {arXiv},
  primaryclass  = {cs.LG},
  url           = {https://arxiv.org/abs/2501.14654}
}

@misc{bigeard2025financeagentbenchmarkbenchmarking,
  title         = {Finance Agent Benchmark: Benchmarking LLMs on Real-world Financial Research Tasks},
  author        = {Antoine Bigeard and Langston Nashold and Rayan Krishnan and Shirley Wu},
  year          = {2025},
  eprint        = {2508.00828},
  archiveprefix = {arXiv},
  primaryclass  = {cs.CE},
  url           = {https://arxiv.org/abs/2508.00828}
}

@misc{opsahlong2026officeqaproenterprisebenchmark,
  title         = {OfficeQA Pro: An Enterprise Benchmark for End-to-End Grounded Reasoning},
  author        = {Krista Opsahl-Ong and Arnav Singhvi and Jasmine Collins and Ivan Zhou and Cindy Wang and Ashutosh Baheti and Owen Oertell and Jacob Portes and Sam Havens and Erich Elsen and Michael Bendersky and Matei Zaharia and Xing Chen},
  year          = {2026},
  eprint        = {2603.08655},
  archiveprefix = {arXiv},
  primaryclass  = {cs.AI},
  url           = {https://arxiv.org/abs/2603.08655}
}

@misc{samuel2025personagymevaluatingpersonaagents,
  title         = {PersonaGym: Evaluating Persona Agents and LLMs},
  author        = {Vinay Samuel and Henry Peng Zou and Yue Zhou and Shreyas Chaudhari and Ashwin Kalyan and Tanmay Rajpurohit and Ameet Deshpande and Karthik Narasimhan and Vishvak Murahari},
  year          = {2025},
  eprint        = {2407.18416},
  archiveprefix = {arXiv},
  primaryclass  = {cs.CL},
  url           = {https://arxiv.org/abs/2407.18416}
}

@inproceedings{xie2024osworld,
  author    = {Tianbao Xie and Danyang Zhang and Jixuan Chen and Xiaochuan Li and Siheng Zhao and Ruisheng Cao and Toh Jing Hua and Zhoujun Cheng and Dongchan Shin and Fangyu Lei and Yitao Liu and Yiheng Xu and Shuyan Zhou and Silvio Savarese and Caiming Xiong and Victor Zhong and Tao Yu},
  title     = {OSWorld: Benchmarking Multimodal Agents for Open-Ended Tasks in Real Computer Environments},
  booktitle = {Advances in Neural Information Processing Systems 38 (NeurIPS 2024)},
  year      = {2024},
  url       = {http://papers.nips.cc/paper_files/paper/2024/hash/5d413e48f84dc61244b6be550f1cd8f5-Abstract-Datasets_and_Benchmarks_Track.html}
}

@misc{lacoste2026cubestandardunifyingagent,
  title         = {CUBE: A Standard for Unifying Agent Benchmarks},
  author        = {Alexandre Lacoste and Nicolas Gontier and Oleh Shliazhko and Aman Jaiswal and Kusha Sareen and Shailesh Nanisetty and Joan Cabezas and Manuel Del Verme and Omar G. Younis and Simone Baratta and Matteo Avalle and Imene Kerboua and Xing Han Lù and Elron Bandel and Michal Shmueli-Scheuer and Asaf Yehudai and Leshem Choshen and Jonathan Lebensold and Sean Hughes and Massimo Caccia and Alexandre Drouin and Siva Reddy and Tao Yu and Yu Su and Graham Neubig and Dawn Song},
  year          = {2026},
  eprint        = {2603.15798},
  archiveprefix = {arXiv},
  primaryclass  = {cs.AI},
  url           = {https://arxiv.org/abs/2603.15798}
}

@misc{dechezelles2025browsergymecosystemwebagent,
  title         = {The BrowserGym Ecosystem for Web Agent Research},
  author        = {Thibault Le Sellier De Chezelles and Maxime Gasse and Alexandre Drouin and Massimo Caccia and Léo Boisvert and Megh Thakkar and Tom Marty and Rim Assouel and Sahar Omidi Shayegan and Lawrence Keunho Jang and Xing Han Lù and Ori Yoran and Dehan Kong and Frank F. Xu and Siva Reddy and Quentin Cappart and Graham Neubig and Ruslan Salakhutdinov and Nicolas Chapados and Alexandre Lacoste},
  year          = {2025},
  eprint        = {2412.05467},
  archiveprefix = {arXiv},
  primaryclass  = {cs.LG},
  url           = {https://arxiv.org/abs/2412.05467}
}

@article{paglieri2024balrog,
  title   = {Benchmarking Agentic LLM and VLM Reasoning On Games},
  author  = {Paglieri, Davide and Cupia{\l}, Bart{\l}omiej and Coward, Sam and Piterbarg, Ulyana and Wo{\l}czyk, Maciej and Khan, Akbir and Pignatelli, Eduardo and Kuci{\'n}ski, {\L}ukasz and Pinto, Lerrel and Fergus, Rob and Foerster, Jakob Nicolaus and Parker-Holder, Jack and Rockt{\"a}schel, Tim},
  journal = {arXiv preprint arXiv:2411.13543},
  year    = {2024}
}

@techreport{anthropic2026opus47,
  author      = {{Anthropic}},
  title       = {Claude {Opus 4.7} System Card},
  institution = {Anthropic},
  year        = {2026},
  month       = apr,
  url         = {https://www.anthropic.com/system-cards},
  note        = {Published April 16, 2026}
}

@misc{openai2026gpt54,
  author       = {{OpenAI}},
  title        = {{GPT-5.4} Thinking System Card},
  year         = {2026},
  month        = mar,
  day          = {5},
  howpublished = {\url{https://openai.com/index/gpt-5-4-thinking-system-card/}},
  note         = {Accessed: 2026-05-03}
}

@techreport{googledeepmind2026gemini31pro,
  author      = {{Google DeepMind}},
  title       = {{Gemini 3.1 Pro} Model Card},
  institution = {Google DeepMind},
  year        = {2026},
  month       = feb,
  url         = {https://deepmind.google/models/model-cards/gemini-3-1-pro/},
  note        = {Published February 2026}
}

@techreport{anthropic2026mythos,
  author      = {{Anthropic}},
  title       = {Claude {Mythos Preview} System Card},
  institution = {Anthropic},
  year        = {2026},
  month       = apr,
  url         = {https://www.anthropic.com/claude-mythos-preview-system-card},
  note        = {Published April 2026}
}

@misc{moonshot2026kimik26,
  author       = {{Moonshot AI}},
  title        = {{Kimi K2.6}: Advancing Open-Source Coding},
  year         = {2026},
  month        = apr,
  howpublished = {\url{https://www.kimi.com/blog/kimi-k2-6}},
  note         = {Moonshot AI blog. Accessed: 2026-05-03}
}

@misc{qwen36plus,
  title  = {{Qwen3.6-Plus}: Towards Real World Agents},
  author = {{Qwen Team}},
  month  = {April},
  year   = {2026},
  url    = {https://qwen.ai/blog?id=qwen3.6}
}

@misc{qwen3.5,
  title  = {{Qwen3.5}: Towards Native Multimodal Agents},
  author = {{Qwen Team}},
  month  = {February},
  year   = {2026},
  url    = {https://qwen.ai/blog?id=qwen3.5}
}

@misc{openai2025gpt51codexmax,
  author       = {{OpenAI}},
  title        = {Building More with {GPT-5.1-Codex-Max}},
  year         = {2025},
  month        = nov,
  howpublished = {\url{https://openai.com/index/gpt-5-1-codex-max/}},
  note         = {Accessed: 2026-05-03}
}

@misc{opencode2026,
  title        = {{OpenCode}: The Open Source AI Coding Agent},
  author       = {{OpenCode Contributors}},
  year         = {2026},
  howpublished = {\url{https://github.com/anomalyco/opencode}},
  note         = {Version 1.14.39; accessed May 5, 2026}
}

@article{deng2025swebenchpro,
  author     = {Xiang Deng and Jeff Da and Edwin Pan and Yannis Yiming He and Charles Ide and Kanak Garg and Niklas Lauffer and Andrew Park and Nitin Pasari and Chetan Rane and Karmini Sampath and Maya Krishnan and Srivatsa Kundurthy and Sean Hendryx and Zifan Wang and Chen Bo Calvin Zhang and Noah Jacobson and Bing Liu and Brad Kenstler},
  title      = {SWE-Bench Pro: Can {AI} Agents Solve Long-Horizon Software Engineering Tasks?},
  journal    = {CoRR},
  volume     = {abs/2509.16941},
  year       = {2025},
  url        = {https://doi.org/10.48550/arXiv.2509.16941},
  eprinttype = {arXiv},
  eprint     = {2509.16941}
}

@inproceedings{li2024deveval,
  author    = {Jia Li and Ge Li and Yunfei Zhao and Yongmin Li and Huanyu Liu and Hao Zhu and Lecheng Wang and Kaibo Liu and Zheng Fang and Lanshen Wang and Jiazheng Ding and Xuanming Zhang and Yuqi Zhu and Yihong Dong and Zhi Jin and Binhua Li and Fei Huang and Yongbin Li and Bin Gu and Mengfei Yang},
  title     = {DevEval: {A} Manually-Annotated Code Generation Benchmark Aligned with Real-World Code Repositories},
  booktitle = {Findings of the Association for Computational Linguistics, {ACL} 2024},
  series    = {Findings of {ACL}},
  pages     = {3603--3614},
  publisher = {Association for Computational Linguistics},
  year      = {2024},
  url       = {https://doi.org/10.18653/v1/2024.findings-acl.214}
}

@inproceedings{yang2024sweagent,
  title     = {{SWE}-agent: Agent-Computer Interfaces Enable Automated Software Engineering},
  author    = {John Yang and Carlos E Jimenez and Alexander Wettig and Kilian Lieret and Shunyu Yao and Karthik R Narasimhan and Ofir Press},
  booktitle = {The Thirty-eighth Annual Conference on Neural Information Processing Systems},
  year      = {2024},
  url       = {https://arxiv.org/abs/2405.15793}
}

@misc{swebenchpro_leaderboard,
  author       = {{Scale AI}},
  title        = {{SWE-Bench Pro} Leaderboard (Public Dataset)},
  howpublished = {\url{https://labs.scale.com/leaderboard/swe_bench_pro_public}},
  year         = {2025},
  note         = {Scale Labs leaderboard. Accessed: 2026-05-03}
}

@misc{anthropic2026claudecode,
  title        = {{Claude Code}: Anthropic's Agentic Coding System},
  author       = {{Anthropic}},
  year         = {2026},
  howpublished = {\url{https://www.anthropic.com/product/claude-code}},
  note         = {Accessed: 2026-05-03}
}

@misc{openai2026codex,
  title        = {{Codex}: OpenAI's Coding Agent},
  author       = {{OpenAI}},
  year         = {2026},
  howpublished = {\url{https://developers.openai.com/codex/cloud}},
  note         = {Accessed: 2026-05-03}
}

@article{merrill2025terminalbench,
  author     = {Mike A. Merrill and Alexander Glenn Shaw and Nicholas Carlini and Boxuan Li and Harsh Raj and Ivan Bercovich and Lin Shi and Jeong Yeon Shin and Thomas Walshe and Estefany Kelly Buchanan and others},
  title      = {Terminal-Bench: Benchmarking Agents on Hard, Realistic Tasks in Command Line Interfaces},
  journal    = {CoRR},
  volume     = {abs/2601.11868},
  year       = {2026},
  url        = {https://doi.org/10.48550/arXiv.2601.11868},
  eprinttype = {arXiv},
  eprint     = {2601.11868}
}

@misc{chan2025mlebenchevaluatingmachinelearning,
  title         = {MLE-bench: Evaluating Machine Learning Agents on Machine Learning Engineering},
  author        = {Jun Shern Chan and Neil Chowdhury and Oliver Jaffe and James Aung and Dane Sherburn and Evan Mays and Giulio Starace and Kevin Liu and Leon Maksin and Tejal Patwardhan and Lilian Weng and Aleksander Mądry},
  year          = {2025},
  eprint        = {2410.07095},
  archiveprefix = {arXiv},
  primaryclass  = {cs.CL},
  url           = {https://arxiv.org/abs/2410.07095}
}

@misc{wang2026cybergymevaluatingaiagents,
  title         = {CyberGym: Evaluating AI Agents' Real-World Cybersecurity Capabilities at Scale},
  author        = {Zhun Wang and Tianneng Shi and Jingxuan He and Matthew Cai and Jialin Zhang and Dawn Song},
  year          = {2026},
  eprint        = {2506.02548},
  archiveprefix = {arXiv},
  primaryclass  = {cs.CR},
  url           = {https://arxiv.org/abs/2506.02548}
}

@misc{chen2025browsecompplusfairtransparentevaluation,
  title         = {BrowseComp-Plus: A More Fair and Transparent Evaluation Benchmark of Deep-Research Agent},
  author        = {Zijian Chen and Xueguang Ma and Shengyao Zhuang and Ping Nie and Kai Zou and Andrew Liu and Joshua Green and Kshama Patel and Ruoxi Meng and Mingyi Su and Sahel Sharifymoghaddam and Yanxi Li and Haoran Hong and Xinyu Shi and Xuye Liu and Nandan Thakur and Crystina Zhang and Luyu Gao and Wenhu Chen and Jimmy Lin},
  year          = {2025},
  eprint        = {2508.06600},
  archiveprefix = {arXiv},
  primaryclass  = {cs.CL},
  url           = {https://arxiv.org/abs/2508.06600}
}

@inproceedings{bandel2026readyforgeneral,
  author    = {Bandel, Elron and Yehudai, Asaf and Shmueli-Scheuer, Michal},
  title     = {Ready For General Agents? Let's Test It.},
  booktitle = {ICLR Blogposts 2026},
  year      = {2026},
  url       = {https://iclr-blogposts.github.io/2026/blog/2026/general-agent-evaluation/}
}

@inproceedings{bandel2026position,
  title     = {Position: Agentic Systems Should be General},
  author    = {Bandel, Elron and Yehudai, Asaf and Lacoste, Alexandre and Ghosh, Avijit and Neubig, Graham and Mitchell, Margaret and Shmueli-Scheuer, Michal and Choshen, Leshem},
  booktitle = {Forty-third International Conference on Machine Learning Position Paper Track},
  year      = {2026},
  url       = {https://openreview.net/forum?id=WMYI7TFDmi}
}

@misc{bandel2026generalagentevaluation,
  title         = {General Agent Evaluation},
  author        = {Elron Bandel and Asaf Yehudai and Lilach Eden and Yehoshua Sagron and Yotam Perlitz and Elad Venezian and Natalia Razinkov and Natan Ergas and Shlomit Shachor Ifergan and Segev Shlomov and Michal Jacovi and Leshem Choshen and Liat Ein-Dor and Yoav Katz and Michal Shmueli-Scheuer},
  year          = {2026},
  eprint        = {2602.22953},
  archiveprefix = {arXiv},
  primaryclass  = {cs.AI},
  url           = {https://arxiv.org/abs/2602.22953}
}

@article{lu2024weblinx,
  title   = {Weblinx: Real-world website navigation with multi-turn dialogue},
  author  = {L{\`u}, Xing Han and Kasner, Zden{\v{e}}k and Reddy, Siva},
  journal = {arXiv preprint arXiv:2402.05930},
  year    = {2024}
}

@inproceedings{zhou2024webarena,
  title     = {Webarena: A realistic web environment for building autonomous agents},
  author    = {Zhou, Shuyan and Xu, Frank F and Zhu, Hao and Zhou, Xuhui and Lo, Robert and Sridhar, Abishek and Cheng, Xianyi and Ou, Tianyue and Bisk, Yonatan and Fried, Daniel and others},
  booktitle = {International Conference on Learning Representations},
  volume    = {2024},
  pages     = {15585--15606},
  year      = {2024}
}

@article{lu2025build,
  title={Build the web for agents, not agents for the web},
  author={L{\`u}, Xing Han and Kamath, Gaurav and Mosbach, Marius and Reddy, Siva},
  journal={arXiv preprint arXiv:2506.10953},
  year={2025},
  url={https://arxiv.org/abs/2506.10953}
}

\appendix
\section{Additional Experiment Results}
\label{app:additional_results}

This section reports supplementary measurements from the main experiments of our coding-agent case study (Section~\ref{sec:casestudy}). We provide them as a reference for downstream cost, efficiency, and behavior analysis.

Table~\ref{tab:main-resource-usage} gives aggregate input tokens, output tokens, and average tool calls per instance for each benchmark-model-harness cell. We report the summed input and output tokens across all runs. The former includes cached tokens, which are often billed at a lower rate. No token count is reported for MSA because the harness does not track or expose such measurements. The tool call count is also reported by the harness after each session concludes. Notably, MSA only features one bash tool, and each harness defines its own set of tools.

\begin{table}[ht]
  \centering
  \caption{Token usage and tool calls for the main experiments. Input tokens include cache reads and cache writes, where reported, and output tokens exclude reasoning tokens. Tool calls are averages per instance over runs. Token counts for MSA are omitted because it does not track token counts.}
  \label{tab:main-resource-usage}
  \small
  \resizebox{\columnwidth}{!}{%
  \begin{tabular}{lllrrr}
    \toprule
    Benchmark & Model & Harness & Input tokens & Output tokens & Tool calls/inst. \\
    \midrule
    DevEval & Claude Opus 4.7 & Claude Code & 114.229M & 6.798M & 9.2 \\
            & GPT-5.4 & Codex & 357.977M & 7.300M & 19.8 \\
            & Gemini 3.1 Pro & OpenCode & 829.617M & 13.624M & 26.6 \\
            & Qwen3.5-397B-A17B & MSA & -- & -- & 37.8 \\
    \addlinespace[0.35em]
    SWE-Bench Pro & Claude Opus 4.7 & Claude Code & 1801.177M & 20.310M & 54.2 \\
                  & GPT-5.4 & Codex & 2078.718M & 17.184M & 63.3 \\
                  & Gemini 3.1 Pro & OpenCode & 1476.167M & 12.190M & 51.5 \\
                  & Qwen3.5-397B-A17B & MSA & -- & -- & 82.1 \\
    \addlinespace[0.35em]
    Terminal-Bench 2.0 & Claude Opus 4.7 & Claude Code & 42.878M & 1.912M & 33.5 \\
                   & GPT-5.4 & Codex & 100.907M & 1.382M & 33.4 \\
                   & Gemini 3.1 Pro & OpenCode & 106.562M & 1.341M & 43.5 \\
                   & Qwen3.5-397B-A17B & MSA & -- & -- & 45.7 \\
    \bottomrule
  \end{tabular}%
  }
\end{table}

\begin{figure*}[t]
    \centering
    \includegraphics[width=0.8\linewidth]{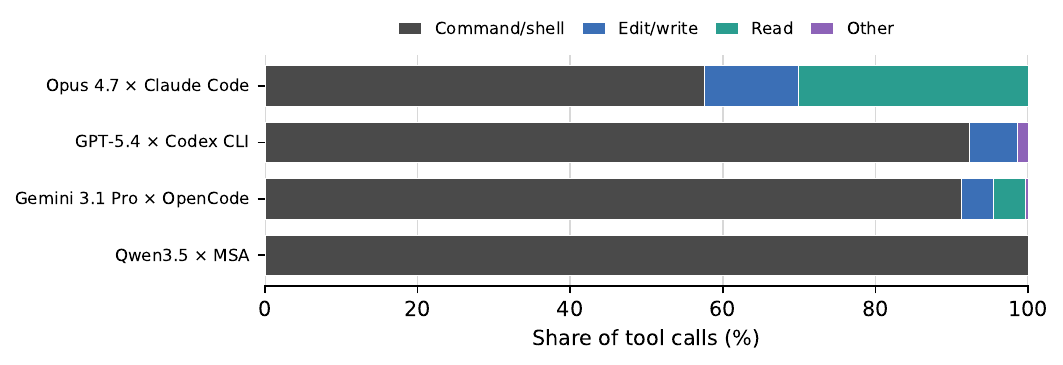}
    \vspace{0.4em}
    \includegraphics[width=0.8\linewidth]{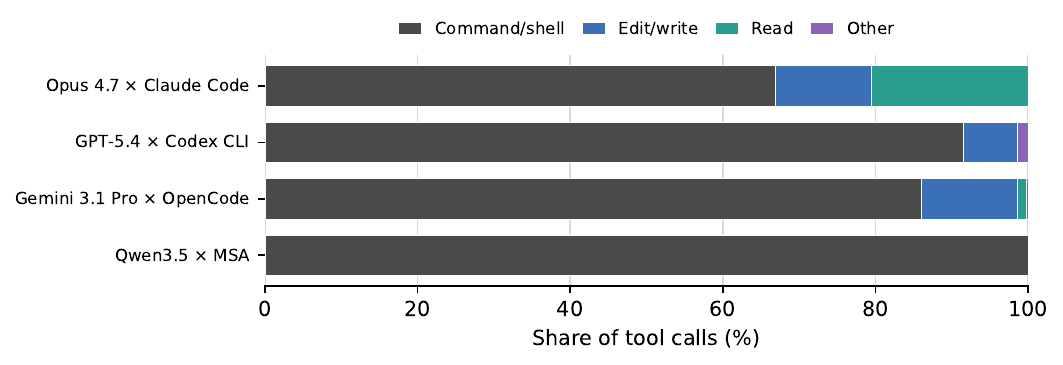}
    \vspace{0.4em}
    \includegraphics[width=0.8\linewidth]{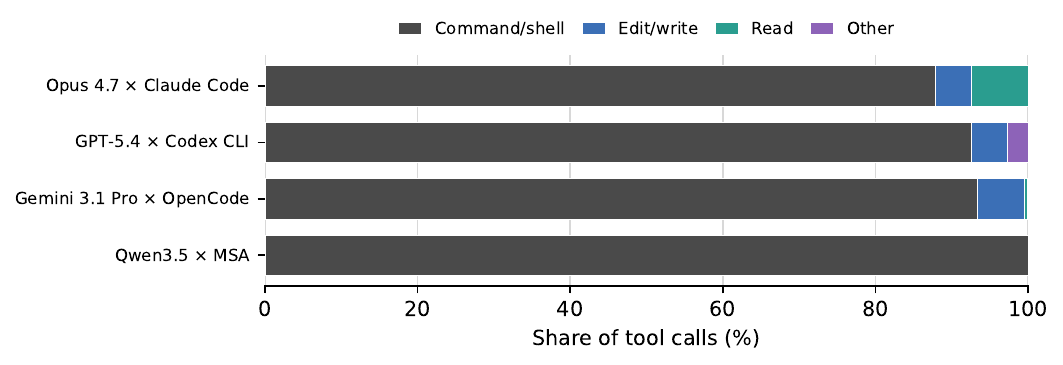}
    \caption{Tool-call distributions for the main experiments, shown separately for DevEval, SWE-Bench Pro, and Terminal-Bench. Each horizontal bar is normalized within a benchmark-model-harness cell. Native harness tool names are mapped into comparable categories: command/shell, edit/write, read, and other.}
    \label{fig:appendix-tool-call-distribution}
\end{figure*}

Figure~\ref{fig:appendix-tool-call-distribution} shows the per-cell distribution of tool calls. We map native tool names into four command categories: command/shell, edit/write, read, and other. The less commonly used tools, including creating a to-do list, planning, and string search, are folded into the other category. However, we observe that the tool call distribution does not reflect the frequency of the implied actions. In particular, the command/shell tool call counts are inflated, as agents often read, edit, and search with shell-native tools instead of relying on the harness-provided function tools.

\begin{figure*}[t]
    \centering
    \includegraphics[width=0.8\linewidth]{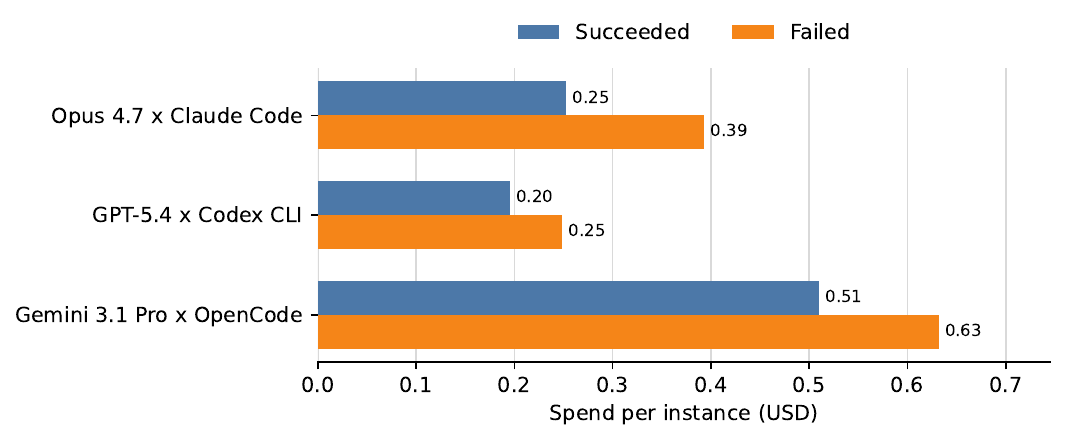}
    \vspace{0.4em}
    \includegraphics[width=0.8\linewidth]{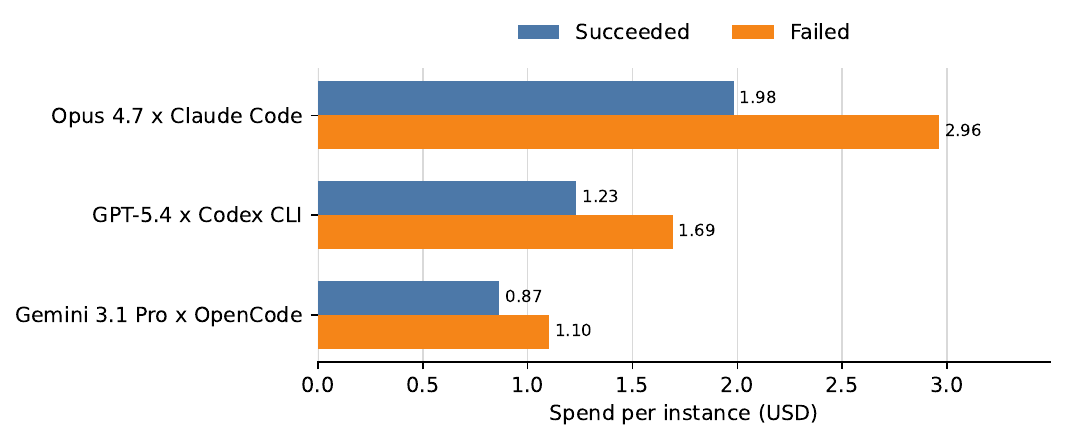}
    \vspace{0.4em}
    \includegraphics[width=0.8\linewidth]{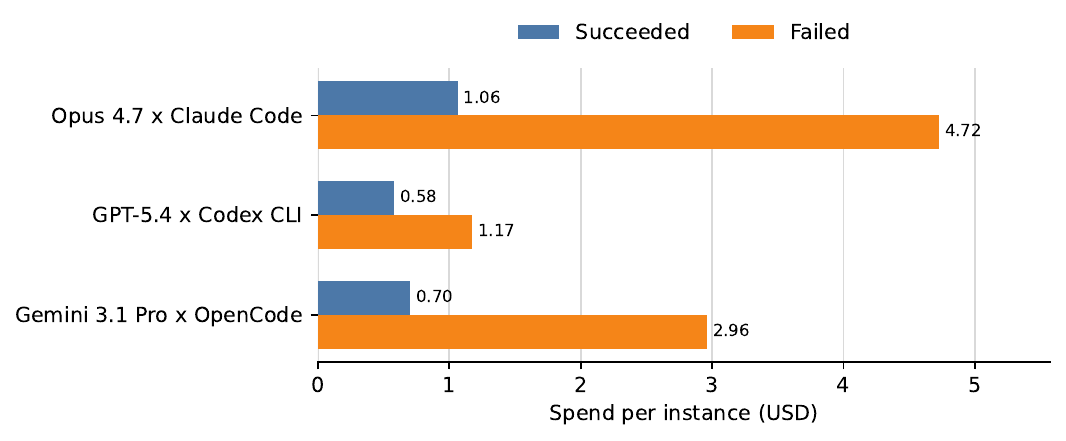}
    \caption{Average spending per instance for succeeded and failed runs in the main experiment. Claude Code costs are harness-reported, while GPT-5.4 and Gemini 3.1 Pro costs are estimated from token usage. MSA is omitted because its logs do not provide reliable spending information.}
    \label{fig:success-failure-cost}
\end{figure*}

\begin{figure*}[t]
    \centering
    \includegraphics[width=0.8\linewidth]{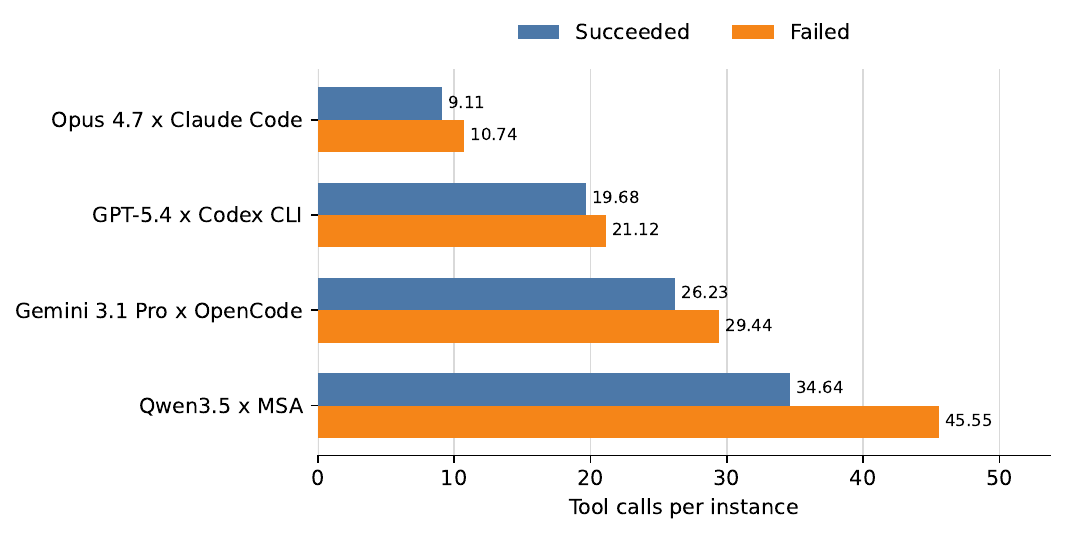}
    \vspace{0.4em}
    \includegraphics[width=0.8\linewidth]{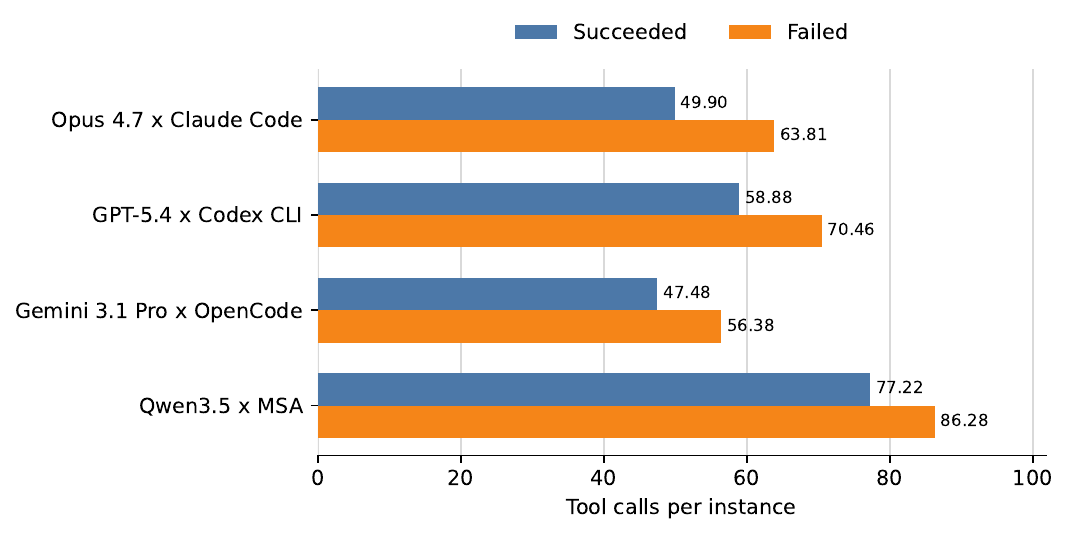}
    \vspace{0.4em}
    \includegraphics[width=0.8\linewidth]{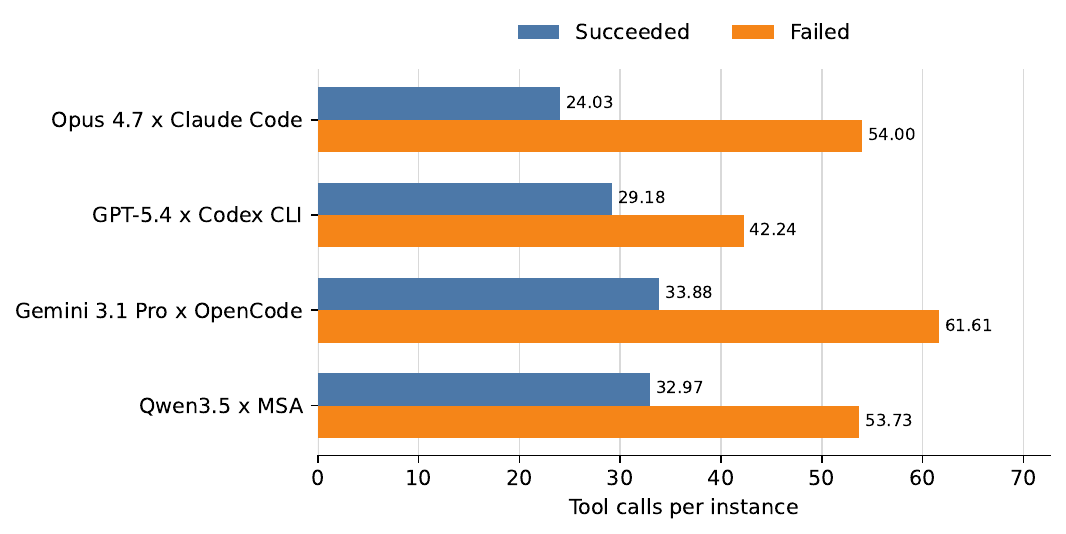}
    \caption{Average tool-call counts per instance for succeeded and failed runs in the main experiment.}
    \label{fig:success-failure-toolcalls}
\end{figure*}

Table~\ref{tab:success-failure-resources}, together with Figures~\ref{fig:success-failure-toolcalls} and~\ref{fig:success-failure-cost}, breaks the tool-call and spending measurements out by whether each run succeeded or failed under the locked scoring rule. We observe an asymmetry of resource spending on the failed and successful instances: the agents consistently make more tool calls and spend more on API costs on the failed instances. This provides indirect evidence that the model recognizes some instances as harder and spends more effort to tackle them.

\begin{table}[ht]
  \centering
  \caption{Average resource use on succeeded and failed main-experiment runs. Claude Code spending is harness-reported, GPT-5.4 and Gemini 3.1 Pro spending are token-rate estimates, and MSA spending is unavailable.}
  \label{tab:success-failure-resources}
  \small
  \resizebox{\columnwidth}{!}{%
  \begin{tabular}{llllrr}
    \toprule
    Benchmark & Model & Harness & Outcome & Tool calls/inst. & Spend/inst. \\
    \midrule
    DevEval & Claude Opus 4.7 & Claude Code & Succeeded & 9.1 & \$0.25 \\
     &  &  & Failed & 10.7 & \$0.39 \\
     & GPT-5.4 & Codex CLI & Succeeded & 19.7 & \$0.20 \\
     &  &  & Failed & 21.1 & \$0.25 \\
     & Gemini 3.1 Pro & OpenCode & Succeeded & 26.2 & \$0.51 \\
     &  &  & Failed & 29.4 & \$0.63 \\
     & Qwen3.5 & MSA & Succeeded & 34.6 & -- \\
     &  &  & Failed & 45.5 & -- \\
    \addlinespace[0.35em]
    SWE-Bench Pro & Claude Opus 4.7 & Claude Code & Succeeded & 49.9 & \$1.98 \\
     &  &  & Failed & 63.8 & \$2.96 \\
     & GPT-5.4 & Codex CLI & Succeeded & 58.9 & \$1.23 \\
     &  &  & Failed & 70.5 & \$1.69 \\
     & Gemini 3.1 Pro & OpenCode & Succeeded & 47.5 & \$0.87 \\
     &  &  & Failed & 56.4 & \$1.10 \\
     & Qwen3.5 & MSA & Succeeded & 77.2 & -- \\
     &  &  & Failed & 86.3 & -- \\
    \addlinespace[0.35em]
    Terminal-Bench & Claude Opus 4.7 & Claude Code & Succeeded & 24.0 & \$1.06 \\
     &  &  & Failed & 17.4 & \$1.23 \\
     & GPT-5.4 & Codex CLI & Succeeded & 29.2 & \$0.58 \\
     &  &  & Failed & 34.6 & \$0.81 \\
     & Gemini 3.1 Pro & OpenCode & Succeeded & 33.9 & \$0.70 \\
     &  &  & Failed & 36.4 & \$0.82 \\
     & Qwen3.5 & MSA & Succeeded & 33.0 & -- \\
     &  &  & Failed & 44.7 & -- \\
    \bottomrule
  \end{tabular}%
  }
\end{table}

Table~\ref{tab:tool-call-failure-rates} details the percentage of failed tool calls issued by each agent. Qwen3.5-397B-A17B $\times$ MSA is not present in the table because the harness executes agent-issued shell commands as they are, without validation. We observe that Gemini 3.1 Pro has the highest error rates on DevEval and Terminal-Bench 2.0, while Claude Opus 4.7 $\times$ Claude Code makes the most tool-call errors on Terminal-Bench 2.0 among the remaining agents.

\begin{table}[t]
  \centering
  \caption{Tool call failure rates in the main experiments.}
  \label{tab:tool-call-failure-rates}
  \small
  \begin{tabular}{lllr}
    \toprule
    Benchmark & Model & Harness & Error Rate \\
    \midrule
    DevEval & Claude Opus 4.7 & Claude Code & 0.12\% \\
            & GPT-5.4 & Codex & 0.04\% \\
            & Gemini 3.1 Pro & OpenCode & 2.44\% \\
    \addlinespace[0.35em]
    SWE-Bench Pro & Claude Opus 4.7 & Claude Code & 0.57\% \\
                  & GPT-5.4 & Codex & 0.20\% \\
                  & Gemini 3.1 Pro & OpenCode & 1.49\% \\
    \addlinespace[0.35em]
    Terminal-Bench 2.0 & Claude Opus 4.7 & Claude Code & 0.64\% \\
                   & GPT-5.4 & Codex & 0.00\% \\
                   & Gemini 3.1 Pro & OpenCode & 0.59\% \\
    \bottomrule
  \end{tabular}
\end{table}

\section{More Case Studies on Agentifying Existing Benchmarks}
\label{app:benchmark-case-studies}

This appendix collects additional case studies showing how existing benchmarks can be adapted to the AAA interaction pattern while preserving their original task and scoring semantics.

\subsection{OSWorld}
\textbf{OSWorld}~\cite{xie2024osworld} (including its continued maintenance in OSWorld-Verified) presents an agent with a task to perform on a provided computer desktop environment and inspects the environment to evaluate whether the task was performed. We adapt the existing Gym-like interface between the benchmark and the agent to run over A2A. We additionally support running the virtualized computer environment in QEMU, which requires less administrative setup than other virtual-machine software and so aligns with AAA's goal of easy-to-launch agents.

\subsection{MLE-bench}
\textbf{MLE-bench}~\cite{chan2025mlebenchevaluatingmachinelearning} presents an agent with a machine-learning problem, including a training dataset and a test dataset, receives outputs for the test set, and evaluates the accuracy of those outputs. MLE-bench launches an agent in a specially prepared container containing machine-learning tools and a copy of the datasets; by contrast, our \greenagent only provides the datasets, receives the predictions over A2A, and leaves the machine-learning computer environment up to the \whiteagent. We also provide a \whiteagent wrapper that receives the datasets and prepares an equivalent container for running an agent compatible with MLE-bench.

\subsection{CyberGym}
\textbf{CyberGym}~\cite{wang2026cybergymevaluatingaiagents} presents an agent with the source code of a vulnerable program and a description of a vulnerability, and asks the agent to provide an input file that triggers the vulnerability. The original implementation sets up the source-code archive and vulnerability description on the machine where the agent works and runs a server that tests the submitted file. Our \greenagent sends the provided files over A2A and receives the submitted file over A2A; again, the \whiteagent brings its own computer environment in which to work. CyberGym packages prebuilt copies of the vulnerable and fixed programs as Docker images, and our \greenagent uses an experimental container-management extension to run these same images.

\subsection{BrowseComp-Plus}
\textbf{BrowseComp-Plus}~\cite{chen2025browsecompplusfairtransparentevaluation} assesses an agent's ability to produce short, precise answers to natural-language research questions, grounded in a fixed corpus of web documents, with an LLM-as-judge grading the answer against a reference. In the original implementation~\cite{chen2025browsecompplusfairtransparentevaluation}, the agent runs as a local process that calls a retrieval backend, typically Pyserini BM25 or dense indexes, before emitting an answer. We move retrieval to the \whiteagent: over A2A, the \greenagent sends a query and the \whiteagent returns the final answer, which lets \whiteagent implementations use their own retrieval methods, whether lexical search, dense indexes, or multi-turn agentic workflows. The original framework downloads the corpus and prebuilt indexes from HuggingFace at runtime and decrypts reference answers on the fly; we instead decrypt queries and reference answers at \greenagent image build time and bundle them into the ground-truth table, so evaluation does not require HuggingFace access. Our reference \whiteagent bundles the BM25 index in its container, and we shard large evaluations across \whiteagent instances, each handling multiple tasks concurrently.




\end{document}